\documentclass[letterpaper, 10 pt, conference]{ieeeconf}  % Comment this line out if you need a4paper

\IEEEoverridecommandlockouts           

\usepackage{math}

\newcommand{\rev}[1]{\textcolor{black}{#1}}

\title{\LARGE \bf DTG : Diffusion-based Trajectory Generation for Mapless Global Navigation}

\author{
Jing Liang$^{1}$, Amirreza Payandeh $^{2}$, Daeun Song$^{1}$, Xuesu Xiao$^{2}$ and Dinesh Manocha$^{1}$ 
\thanks{$^{1}$University of Maryland, College Park. $^{2}$George Mason University}
}

\begin{document}
\maketitle
\thispagestyle{empty}
\pagestyle{empty}

\begin{abstract}
We present a novel end-to-end diffusion-based trajectory generation method, DTG, for mapless global navigation in challenging outdoor scenarios with occlusions and unstructured off-road features like grass, buildings, bushes, etc. Given a distant goal, our approach computes a trajectory that satisfies the following goals: (1) minimize the travel distance to the goal; (2) maximize the traversability by choosing paths that do not lie in undesirable areas. Specifically, we present a novel Conditional RNN(CRNN) for diffusion models to efficiently generate trajectories. Furthermore, we propose an adaptive training method that ensures that the diffusion model generates more traversable trajectories. We evaluate our methods in various outdoor scenes and compare the performance with other global navigation algorithms on a Husky robot. In practice, we observe at least a \textbf{$15\%$} improvement in traveling distance and around a $7\%$ improvement in traversability.
Video and Code: \url{https://github.com/jingGM/DTG.git}.\end{abstract}

\section{Introduction}
\label{sec:intro}

Global navigation is used to compute the trajectories of robots in large-scale environments~\cite{ganesan2022global, gao2019global, ort2018autonomous}. Global navigation is widely used in various tasks, such as autonomous driving~\cite{ort2018autonomous, yurtsever2020survey}, last-mile delivery~\cite{hoffmann2018regulatory, chen2021adoption}, and search and rescue operations~\cite{davids2002urban}. However, various challenges have to be addressed to successfully conduct global navigation tasks~\cite{huang2005online, giovannangeli2006robust, shah2022viking, liang2023mtg}.

Mapless Navigation is Critical for Outdoor Global Navigation. To facilitate planning, a global map is computed for many global navigation strategies~\cite{ganesan2022global, gao2019global, psotka2023global}. However, the acquisition of an accurate and detailed map poses significant challenges, particularly for outdoor navigation tasks with frequently changing environments~\cite{wijayathunga2023challenges} due to weather changes~\cite{zhang2020virtual, ort2020autonomous}, temporary construction sites~\cite{jeong2021motion}, and hazardous areas~\cite{eom2010hazardous}. Therefore, it is important to develop mapless navigation strategies for general outdoor scenes. Nevertheless, there are various challenges associated with mapless outdoor navigation, including traversability analysis, optimality assurance, constraints satisfaction, etc.~\cite{
% giovannangeli2006robust, 
shah2022viking, lavalle2006planning}.

\begin{figure}
    \centering
    \includegraphics[width=0.9\linewidth]{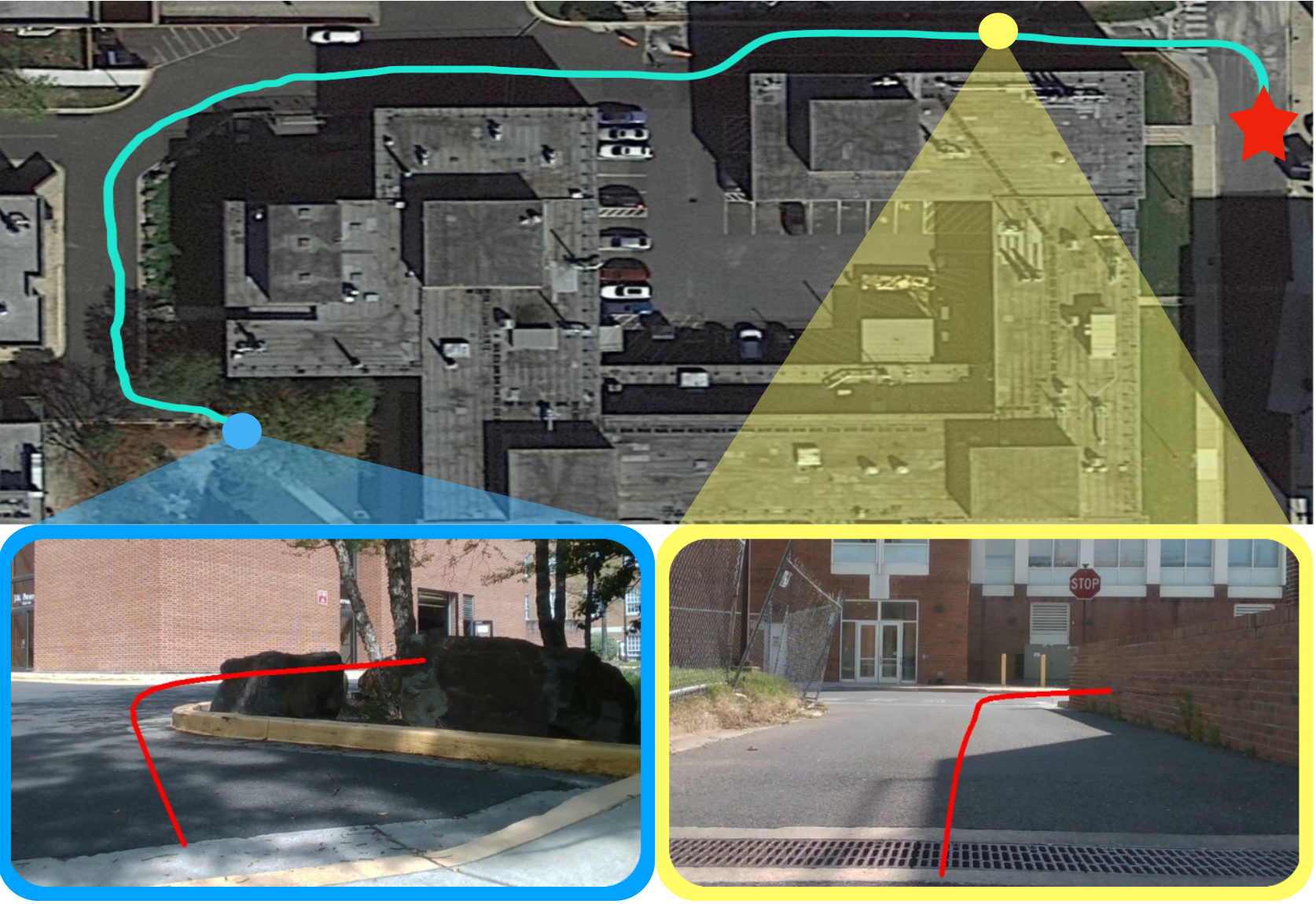}
    \caption{DTG generates trajectories in traversable areas with the shortest travel distance to the target (red star). The blue and yellow boxes show the efficacy of the generated trajectories in scenarios including bushes and buildings.}
    \label{fig:front}
    \vspace{-2em}
\end{figure}

\textit{Computing Traversable Trajectories for Robotic Navigation:} In complex outdoor scenarios, traversability analysis is critical for safe outdoor navigation because outdoor environments contain various challenging terrains, plants, buildings, trees, etc.~\cite{liang2022adaptiveon, weerakoon2023graspe, xiao2021learning, karnan2022vi, pokhrel2024cahsor}. Traditionally, perception and planning are decoupled as two different tasks~\cite{canny1988complexity,manocha1992algebraic,lavalle2006planning}, perception to detect traversable areas and planning to generate waypoints. However, the traversable maps created by these approaches may not be the most effective representation of the environment. For example, in learning-based approaches~\cite{liang2021crowd, sridhar2023nomad, shah2023vint}, the high-dimensional observations are usually processed into a vector to encode the traversability information. Furthermore, for mapless navigation with occlusions, estimating traversable areas behind occluding objects can be challenging. Therefore, effectively modeling the complex traversability information in outdoor scenarios is important.

\textit{Optimal Trajectory Generation Towards Designated Goals.} Beyond ensuring traversability, navigation problems are often formulated as optimization problems~\cite{liang2021vo, liang2022adaptiveon, malyuta2022convex}, aiming to compute optimal trajectories under constraints. Some of the common optimality criteria include minimizing the running time or the traveling distance~\cite{liang2021crowd, malyuta2022convex, gasparetto2015path, sanchez2021path}. However, in mapless navigation, if the goal is far away, the absence of an exact map complicates the evaluation of the remaining travel distance, e.g., whether traversing through a wider or narrower passage will result in shorter travel distance~\cite{wu2022image, jang2021hindsight}.

\textit{Generating Trajectories with Traversability and Optimality.} Learning-based approaches demonstrate remarkable performance in outdoor navigation tasks~\cite{shah2022viking, sridhar2023nomad, liang2022adaptiveon}. However, the selection of an effective model to accurately generate trajectories that are optimal in terms of traveling distance and to satisfy traversability constraints is challenging. While on-policy reinforcement learning-based outdoor navigation approaches~\cite{liang2022adaptiveon, doukhi2022deep} can generate optimal trajectories, it is not safe to try failure cases, such as collision or flipping over, in complex outdoor scenarios. Many approaches apply supervised learning with different models~\cite{kahn2021land, shah2023vint, sridhar2023nomad, liang2023mtg}. The diffusion model shows promising performance in different robotic applications, such as picking and pushing operations of robotic arms~\cite{chi2023diffusion} and navigation~\cite{sridhar2023nomad}. However, the denoising procedure with U-Net~\cite{chi2023diffusion} is still not computationally efficient for real-time navigation. However, NoMaD still requires very close subgoal images, which is not fully mapless, and the nearest subgoals need to be very close to the robot. Furthermore, the diffusion models are only trained to imitate the ground truth without any additional constraints, such as the traversability constraints.

\textbf{Main Innovations: } We introduce the diffusion mechanism in the mapless outdoor global navigation task and present a novel end-to-end approach, DTG, to generate trajectories to alleviate the challenges of complex outdoor mapless environments. Under the condition of the environmental information, the diffusion model takes a random Gaussian noise and denoises it in multiple steps to predict a traversable trajectory with short travel distance to the goal. We also demonstrate the benefits of our approach in complex outdoor scenarios with occlusions and unstructured elements. The major contributions include:
\begin{enumerate}
    \item \textbf{A Novel End-to-end Diffusion-based Trajectory Generator for Global Navigation: } We apply diffusion models in the mapless outdoor global navigation task. The diffusion-based generator generates the trajectories with decent traversability and short future travel distances in the global navigation task with a distant goal ($>$50 meters).
    \item \textbf{A Novel Conditional RNN (CRNN) Model for The Diffusion Model:} To make diffusion models run in real time for navigation, we propose CRNN, which takes the environment information as conditions and generates trajectories in real time for global navigation.
    \item \textbf{Adaptive Training to Enhance Traversability:} To enhance the traversability of the generated trajectories, we propose a new method to adaptively apply traversability loss to different diffusion steps according to the historic loss of the steps.
    % train the diffusion-based trajectory generation. The traversability loss is adaptively applied to 
    \item \textbf{Performance Improvement in Global Navigation:} We demonstrate the benefits of our approach, DTG, in complex outdoor scenarios with occlusions and complex features, such as bushes, grass, and other off-road, non-traversable areas. We compare with the state-of-the-art trajectory generation approaches (ViNT~\cite{shah2023vint}, NoMaD~\cite{sridhar2023nomad}, and MTG~\cite{liang2023mtg}) for outdoor global navigation and observe at least a {$\mathbf{15\%}$} improvement in future travel distance and a $7\%$ improvement in traversability. We also qualitatively show the benefits of our approach: around corners with occlusions DTG generates trajectories with better traversability, and around narrow spaces it can generate trajectories with shorter path lengths. Our proposed CRNN also achieves traversability and travel distances comparable to U-Net~\cite{chi2023diffusion} with less running time and smaller a model size for real-time navigation.
\end{enumerate}

\section{Prior Work and Background}
\label{sec:background}
In this section, we review the related works on trajectory generation in challenging outdoor navigation.

% outdoor navigation
\textbf{Outdoor Navigation: } Navigating robots in outdoor environments presents significant challenges due to occlusions, weather conditions, construction sites, and diverse and complex terrains like grass, bushes, mud, and sharp elevation changes~\cite{weerakoon2023graspe, xiao2021learning, karnan2022vi, pokhrel2024cahsor,canny1988complexity,manocha1992algebraic}. Various strategies~\cite{liang2022adaptiveon, weerakoon2023graspe, xu2021applr, xu2021machine} have been proposed to address these issues. Motion planning techniques aim for stable and safe robot movement across different terrains~\cite{liang2022adaptiveon, weerakoon2023graspe}, using adaptive approach~\cite{liang2022adaptiveon} and reinforcement learning~\cite{ xu2021applr, xu2021machine} to train neural networks for waypoint generation. Global planning requires a comprehensive cost map for path planning~\cite{fusic2021optimal, li2021openstreetmap} and accurate robot localization~\cite{akai2017robust, lowry2015visual} to follow the paths. However, those map-based navigation approaches can be computationally expensive and require a significant amount of overhead to maintain the maps. To solve this issue, instead of maintaining a comprehensive map, Sridhar et al.~\cite{shah2023vint, sridhar2023nomad} and Hirose et al.~\cite{hirose2023exaug} propose generating topological maps and using images as subgoals for navigation, but those approaches still require initial runs in the environment to gather subgoal images. Mapless navigation techniques are used to navigate without relying on maps. Giovannangeli et al.\cite{giovannangeli2006robust} and Liang et al.~\cite{liang2022adaptiveon} generate actions in a mapless manner, but they focus on local planning without addressing long-distance navigation. MTG~\cite{liang2023mtg} represents a step forward by providing long-distance navigation trajectories, yet it doesn't optimize for the best path. In contrast, we propose a novel approach, DTG, which generates traversable trajectories with short travel distances towards a distant goal in large-scale outdoor settings without a map.

% traversability analysis
\textbf{Traversability Analysis: } Traversability analysis plays a critical role in robot navigation, distinguishing between navigable and non-navigable areas within an environment. This task is often handled by separating perception and planning~\cite{fan2021step, 
% sikand2022visual
gasparino2022wayfast, jonasfast}, utilizing various sensors to assess the terrain. Cameras (RGB or RGB-D) are commonly employed to analyze the terrain~\cite{hosseinpoor2021traversability, jonasfast}, enabling segmentation and the creation of cost maps for navigation. Similarly, Lidar sensors are used for their ability to generate elevation maps through geometric information~\cite{xue2023traversability, Fankhauser2018ProbabilisticTerrainMapping}. However, reliance on cost or elevation maps, while visually intuitive for humans, poses computational costs for robots due to the extra processing and encoding of the maps. To address this, end-to-end learning approaches~\cite{xiao2022motion, liang2022adaptiveon, liang2023mtg} have been developed to encode environmental information directly into neural networks, thereby streamlining the navigation process. In our work,  we also apply an end-to-end learning approach to encode the observation information.

% trajectory generation
\textbf{Trajectory Generation: } Trajectory generation in outdoor scenarios varies significantly for different applications. Some methods are designed specifically for autonomous driving~\cite{varadarajan2022multipath, nayakanti2023wayformer, shi2022motion} with LSTM~\cite{altche2017lstm} or Gaussian Mixture Models~\cite{varadarajan2022multipath} to predict trajectories based on historical movements. Smaller robots' trajectory generation strategies~\cite{kahn2021land, shah2022viking, shah2023vint} leverage Bayesian-based methods~\cite{liang2023mtg, yang2020online}, GANs (Generative Adversarial Networks)~\cite{gupta2018social,10323465 } etc to compute feasible paths through complex environments. 
% This approach specifically addresses the challenges of navigating small robots in a large-scale mapless environment. 
Existing methods for small robots global navigation, such as ViNT~\cite{shah2023vint} and NoMaD~\cite{sridhar2023nomad}, rely on comparing current images to pre-recorded subgoal images to navigate, but they cannot recognize the perceived images with significant differences from the subgoal images or in completely unknown environments, so sophisticated choices of subgoals are necessary~\cite{shah2023vint}. Moreover, they do not consider the optimality of the trajectories in terms of travel distance to the goals and require prior knowledge (subgoal images) of the environment, limiting their applicability to unknown or dynamically changing areas. MTG~\cite{liang2023mtg} uses CVAE~\cite{cvae} to generate trajectories in traversable areas, but it doesn't consider the optimality of the trajectories. Diffusion models~\cite{ho2020denoising, song2020denoising} have been used for robotics applications, such as picking and pushing objects~\cite{chi2023diffusion} and navigation~\cite{sridhar2023nomad}. In our approach, we propose a novel diffusion-based mapless trajectory generator to generate optimal trajectories with short travel distances to goals and train the generation with traversabilities metrics. 

\section{Approach}
\label{sec:approach}

In this section, we formulate the problem of mapless global navigation and describe how our approach, DTG, addresses this problem.

\begin{figure*}
\centering
\includegraphics[width=0.8\linewidth]{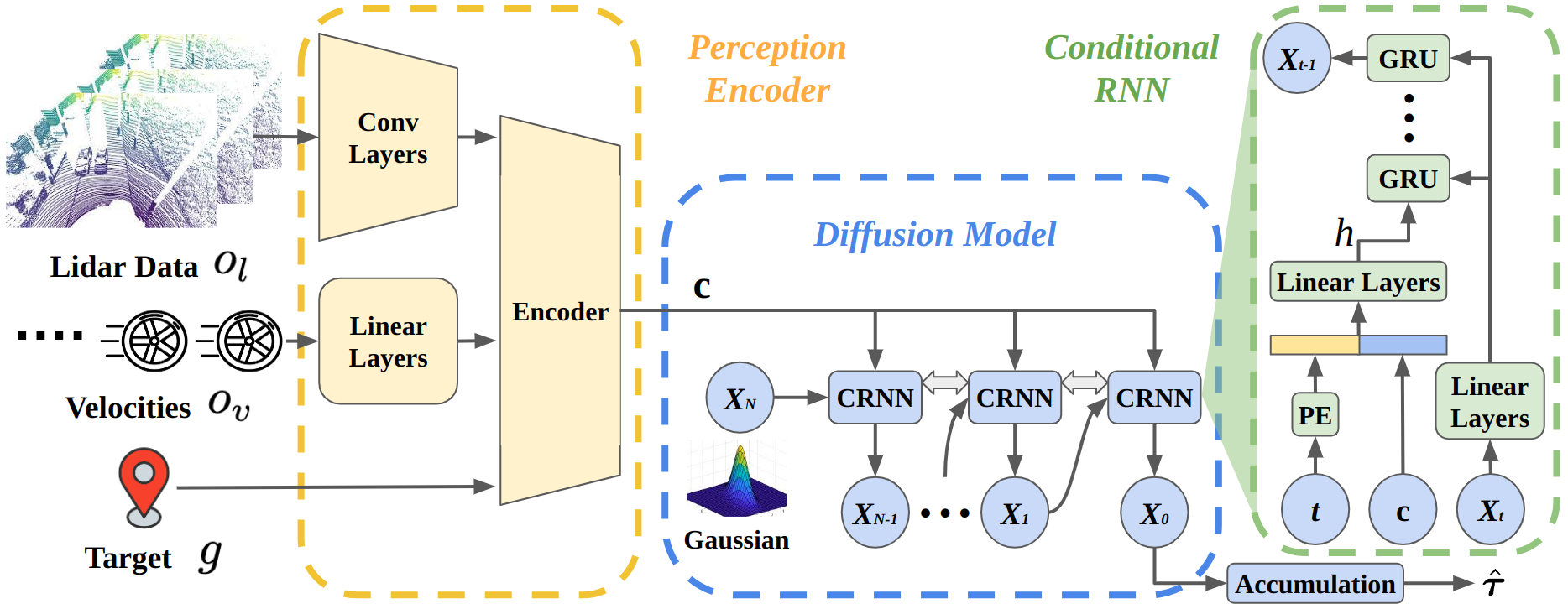}
\caption{\textbf{DTG Architecture:} DTG has two models: \textbf{Perception Encoder} and \textbf{Diffusion Model}. Perception Encoder encodes the observation information, $\o = \set{o_l, o_v , g}$, to the condition vector, $\c$. The Diffusion Model takes a Gaussian distribution to generate a trajectory $\hat{\btau}$ under the condition $\c$.}
    \label{fig:architecture}
\vspace{-2em}
\end{figure*}

\subsection{Problem Definition}
\label{sec:problem_definition}
The problem we are solving is outdoor mapless global navigation. Given a  distant goal $g\in\co_g$
% , which is assumed to be mostly out of the view of a robot 
in a large-scale environment, our model generates trajectories in the robot's traversable areas while trying to minimize the travel distance to the goal. We assume the trajectory generator ignores small and dynamic obstacles, which can be handled by the local planners.

Perceptual sensors used in our approach include a 3D LiDAR and the robot's odometer. We utilize $C_l$ consecutive frames of LiDAR perception, $\co_l$, to capture the static and dynamic information of the environment in the robot's vicinity. The odometer provides the latest $C_v$ consecutive frames of the robot's velocities, $\co_v$, to encode the robot's dynamic status. In this global navigation task, the goal is denoted as $g = \set{g_x, g_y} \in \co_g$. Thus, the observation contains $\co = \set{\co_l, \co_v, \co_g}$. Given the observation $\co$, our trajectory generator $DTG_\theta$ generates a trajectory $\btau=\{\w_1,..., \w_M\}$ that both satisfies traversability constraints and has minimum travel distance to the goal $g$. Here, $\w_m = \{x_m, y_m\}$ represents the waypoints in the generated trajectory, which contains M waypoints in total starting from the robot. We define the travel distance of the trajectory $\btau$ as the length of the shortest path from the last waypoint $\w_M$ of $\btau$ to the target $g$. For each observation $o\in \co$, we have $DTG_\theta(\o) = \btau$. The problem can be formulated as the following equation:

\begin{align}
    \hat{\theta} = \argmin{\theta} \left( h(DTG_\theta(o)_l, g) + \beta f(DTG_\theta(o), \Tilde{\ca}) \right)
\end{align}
where $h(\cdot,\cdot)$ represents the travel-distance function between the last waypoint $\w_M$ and the goal $g$. $\ca$ is the traversable area around the robot and $\Tilde{\ca}$ represents non-traversable areas. $\beta$ is a hyperparameter and $f(\cdot, \cdot)$ calculates the traversability of the trajectory $\btau$, which is the portion of non-traversable areas covered by the trajectory $\btau$. Therefore, the function achieves two targets: 1. {Optimize the travel distance}; and 2. {Satisfy traversability constraints}. The inference of our approach only takes $\o\in \co$ as input, but to train the models, we require $\ca$ to calculate the traversability ground truth.

For training, DTG uses a similar traversability map as in MTG~\cite{liang2023mtg}, where the off-road areas and buildings are not traversable, but sidewalks, pavements, and drivable roads are traversable areas. The travel-distance function $h(\cdot,\cdot)$ can be handled by the A* algorithm, starting from the last waypoint $\w_M$ to the goal $g$. The travel distance is the length of the calculated A* path. The non-traversable area $\Tilde{\ca}$ can also be directly extracted from the traversability map. 
%Then, the problem becomes optimizing the travel distance and traversability of the generated trajectory.  

Since we already have the traversability map and use path planning methods to calculate the travel distance of trajectories, we can directly use the trajectories as the ground truth of our model, DTG, in the training. 
Thus, we redefine the problem as Equation \ref{eq:problem_definition}.

\begin{align}
    \hat{\theta} = \argmin{\theta} \left( d(DTG_\theta(o), \btau_{gt})  + \beta f(DTG_\theta(o), \Tilde{\ca}) \right),
    \label{eq:problem_definition}
\end{align}
where $\btau_{gt}$ is the ground truth trajectory with the shortest travel distance to the goal $g$ and lies in traversable areas. $d(\cdot,\cdot)$ calculates the distance between the generated trajectory and the ground truth trajectory $\btau_{gt}$. Since the ground truth and inputs are all given, this problem can be handled by a supervised learning method.

\subsection{Architecture}
Figure~\ref{fig:architecture} shows the end-to-end architecture of our approach, DTG. There are two models in the pipeline: \textbf{Perception Encoder} $P_\theta(\cdot)$ and \textbf{Diffusion Model} $D_\theta(\cdot)$. 
Here we denote all model parameters as $\theta$. Details about the layer configurations are in Appendix~\ref{appdenix}\cite{supplements}.

\subsubsection{\textbf{Perception Encoder}}
\label{sec:perception}
The Perception Encoder encodes the LiDAR, Velocities, and Target information into a vector as the condition input of the Diffusion Model:
\begin{align}
    \c = P_{\theta}(o_l, o_v, g) = p^e_{\theta}(p^c_{\theta}(o_l), p^v_{\theta}(o_v), g),
\end{align}
where $o_l\in\co_l$, $o_v\in\co_v$ and $g\in\co_g$. $p^c_\theta(\cdot)$ represents the model PointCNN~\cite{hua2018pointwise}. $p^v_\theta(\cdot)$ is a sequence of Linear layers to process the robot's dynamic status with historic velocities. As shown in Figure \ref{fig:architecture}, the Encoder layer takes the concatenated embeddings from $p^c_{\theta}(o_l), p^v_{\theta}(o_v)$, and $g$ and encodes the embeddings to a vector, $\c$, as the condition of the Diffusion model. The Encoder, $p^e_\theta(\cdot)$, also composes a sequence of Linear layers.

\subsubsection{\textbf{Diffusion Model}}
\label{sec:diffusion}
The Diffusion model generates high-quality data by progressively denoising a Gaussian noise to some target data~\cite{ho2020denoising, song2020denoising}, and it shows promising capabilities in robotics tasks~\cite{chi2023diffusion}. In our approach, the diffusion model takes the conditional vector $\c$ from the Perception Encoder for each observation and denoise a Gaussian distribution to a trajectory $\hat{\btau}$. In recent years, multiple diffusion training strategies have been proposed, including DDPM~\cite{ho2020denoising}, DDIM~\cite{song2020denoising}, and VDM~\cite{vdm}. Because VDM~\cite{vdm} shows higher quality in data generation and better convergence than others, in this approach, we use the VDM~\cite{vdm} strategy to generate trajectories. In the diffusion process, neural networks (diffusion cells) are trained to learn noise or noisy data, and these two procedures are mathematically equivalent. Because of faster convergence and smaller converged loss, we use the diffusion cells to predict trajectories in our approach, DTG. Our diffusion generator contains N steps. As shown in Figure \ref{fig:architecture}, from the final output ($\x_0$) to the Gaussian noise ($\x_N$), noise is introduced in each step, resulting in a noisy trajectory denoted as $\Tilde{\x}_t = \sqrt{\alpha_t} \x_{t-1} + \sqrt{1-\alpha_t} \epsilon$, where $\alpha_t$ is the noise ratio at step $t$, which ranges from 1 to N. $\epsilon$ represents the noise itself. The predicted trajectory in step $t$ is $\hat{\x}_{t-1}=R_\theta(\Tilde{\x}_t, t)$, where $R_\theta(\cdot)$ is a diffusion cell. Because the U-Net~\cite{ho2020denoising, chi2023diffusion} is very computationally intensive, we propose a novel diffusion cell to reduce the computational cost in Figure~\ref{fig:architecture}. Since we need the generated trajectories to remain always within traversable areas, we require the diffusion cell to also integrate environmental information. Inspired by~\cite{chi2023diffusion}, we propose a Conditional RNN (CRNN), shown on the right side of Figure~\ref{fig:architecture}, that takes environmental information vector $\c$ and the step number $t$ as conditions for the generator, denoted by $R_\theta(\Tilde{\x}_{t-1}, t, \c)$. Thus, we have the CRNN cell:
\begin{align}
d^h_\theta(t,\c) &=f^2_\theta(f^1_\theta(d^p(t)),\c)=\h,\\
    r^k_\theta(\Tilde{\x}_t, \h) &= d^g_\theta(\Tilde{\x}_t, \h), \\
    R_\theta(\Tilde{\x}_t, t,\c) &= r^1_\theta( \Tilde{\x}_t, ... r^1_\theta(\Tilde{\x}_t, \h)),
\end{align}
where $d^h_\theta(\cdot)$ calculates the hidden vector $\h$ for GRU cells. $d^p(\cdot)$ represents Sinusoidal positional embedding. $f^1_\theta(\cdot)$ and $f^2_\theta(\cdot)$ are all Linear layers. $k\in \set{1,..., K}$ represents the steps in the CRNN model. Each step is $r^k_\theta(\cdot)$. $d^g_\theta(\cdot)$ is the GRU cell and $\h$ is the hidden vector for the GRU cell. We also compare with U-Net~\cite{chi2023diffusion} as the conditional encoder and show the results in Table~\ref{tab:ros_quanti} demonstrating that we have comparable results, in terms of traversability and travel distance, but significantly less computational cost.

Then we have the sampled trajectory:
\begin{align}
    \hat{\x}_t = R_\theta(...R_\theta(\bxi, N, \c), 1, \c),
\end{align}
where $\bxi$ is sampled from a random Gaussian distribution, $\cn(\mu,\nu)$. As shown in Figure~\ref{fig:architecture}, all CRNN models share the same parameters. The output of the diffusion model $\hat{\x}_0$ composes a sequence of $\set{\Delta x_m, \Delta y_m}$, and the waypoint positions $w_m = \set{x_m, y_m}\in \hat{\btau}$ are calculated by accumulating the incremental distances.

\subsection{Training Strategy}
According to the problem defined in Section~\ref{sec:problem_definition}, we have two targets to achieve: 1. Reduce the distance between the generated trajectories and the ground truths, and 2. Minimize the portion of generated trajectories in non-traversable areas. Since our approach is an end-to-end model, we jointly train both targets with an adaptive training strategy:

\subsubsection{Train the generated trajectories to align with the ground truth paths, which have the shortest travel distance}
As mentioned in Section~\ref{sec:diffusion}, our approach uses the training of predicting trajectories. According to the diffusion loss in~\cite{vdm}, we formulate our diffusion loss as
\begin{align}
    \cl_d = \expect{t,\epsilon_d} \left( (\text{SNR}(t-1) - \text{SNR}(t)) \norm{\hat{\btau}_t - \btau_{gt}}^2_2 \right),
\end{align}
where $t\in\set{1,...,N}$ and $\epsilon_d \in \cn(0, \I)$. $\btau_{gt}$ is the ground truth trajectory because optimizing VDM~\cite{vdm} boils down to predicting the original ground truth. SNR$(t) = \frac{\Bar{\alpha}_t}{1- \Bar{\alpha}_t}$. $\Bar{\alpha}_t=\prod_{i=1}^t \alpha_i$ is defined the same as in \cite{ho2020denoising}. To simplify the loss function, because SNR(t) is monotonically decreasing, $(\text{SNR}(t-1) - \text{SNR}(t))$ is always positive. We find similar training results after removing this term during training. Thus, the loss function is simplified as:
\begin{align}
    \cl_d = \expect{t,\epsilon_d} \norm{\hat{\btau}_t - \btau_{gt} 
    % s(\btau_{gt}, t)
    }^2_2.
\end{align}

\subsubsection{{Adaptive Training of Diffusion Models}}

The current loss function only trains the diffusion model to imitate the ground truth trajectories, and we still need to add constraints to train the generated trajectories only in traversable areas. However, because the diffusion model denoises the Gaussian distribution step by step, we cannot directly add the traversability constraint in each step or the training gets sabotaged. Thus, we propose an adaptive strategy to add the traversability constraints.

\begin{align}
    \cl_t &= \expect{t\in \b_t} \exp{\left(1 - \frac{1}{M}\sum_{m=1}^M\min{d(\Tilde{\ca}, w_m)}\right)} ,
    % \\
    % \cl_t &=  l_t(\hat{\x}_t, \Tilde{\ca})
    \label{eq:traversability_loss}
\end{align}
where $d(\cdot)$ calculates the distance between the current waypoint $w_m$ in $\hat{\btau}_t$ and its nearest non-traversable area The distance function clips the value to $[0-1]$ meters. We sample t from the diffusion step buffer $\b_t$, which stores the currently available diffusion steps. As shown in Algorithm \ref{alg:adaptive_schedule}, $m_\theta(\cdot)$ is the neural network model and $\theta$ represents the parameters. $i$ is the index of training epochs and $N_t=10$ is the maximum step number to apply to the traversability loss. $l_d(t-1)$ calculate the average of $5$ last recorded diffusion loss values at step $t-1$. If the average loss value is smaller than the threshold $H_d$, we add this step to the step buffer $\b_t$. We increase the range of diffusion steps incrementally to adapt the training of traversability to the training of ground truth trajectories. Finally, we have the total loss function for DTG: $\cl = \beta_1\cl_d + \beta_2 \cl_t$, where $\beta$ represents hyperparameters.

\begin{algorithm}
\caption{Adaptive training schedule of DTG: adaptively apply the traversability loss to DTG.}\label{alg:adaptive_schedule}
\begin{algorithmic}
\REQUIRE $N_t \gets 10$
\REQUIRE $\b_t = \set{\;}$

\FOR{$i \gets 0$ to Total Epochs}
    \STATE $t = \text{RandomSample}(0, N-1)$
    \STATE $\hat{\x}_t = m_\theta (\o, t)$
    \IF{$i\geq 1$ and $t< N_t$ and $l_d(t-1)< H_d$}
    \STATE Add $t-1$ to $\b_t$
    \ENDIF
    \IF{$\abs{\b_t}>0$ and $t\in \b_t$}
    \STATE $\cl = \beta_1 \cl_d(\hat{\x}_t, \btau_{gt}, t) + \beta_2 \cl_t(\hat{\x}_t, \Tilde{A})$
    \ELSE
    \STATE $\cl = \cl_d(\hat{\x}_t, \btau_{gt})$
    \ENDIF
\ENDFOR
\end{algorithmic}
\end{algorithm}
\vspace{-1em}

\begin{figure*}[!ht]
  \centering
  \begin{tabular}{ | c | c | c | c |}
    \hline
   % \multicolumn{3}{|c}{\textbf{Traversability Analysis}} & \multicolumn{3}{|c|}{\textbf{Heuristic Analysis}} \\ \hline
   \textbf{DTG} & \textbf{MTG}~\cite{liang2023mtg} & \textbf{NoMaD}~\cite{sridhar2023nomad} & \textbf{ViNT}~\cite{shah2023vint}  \\ \hline
   
    \begin{minipage}{.22\linewidth} \includegraphics[width=\linewidth]{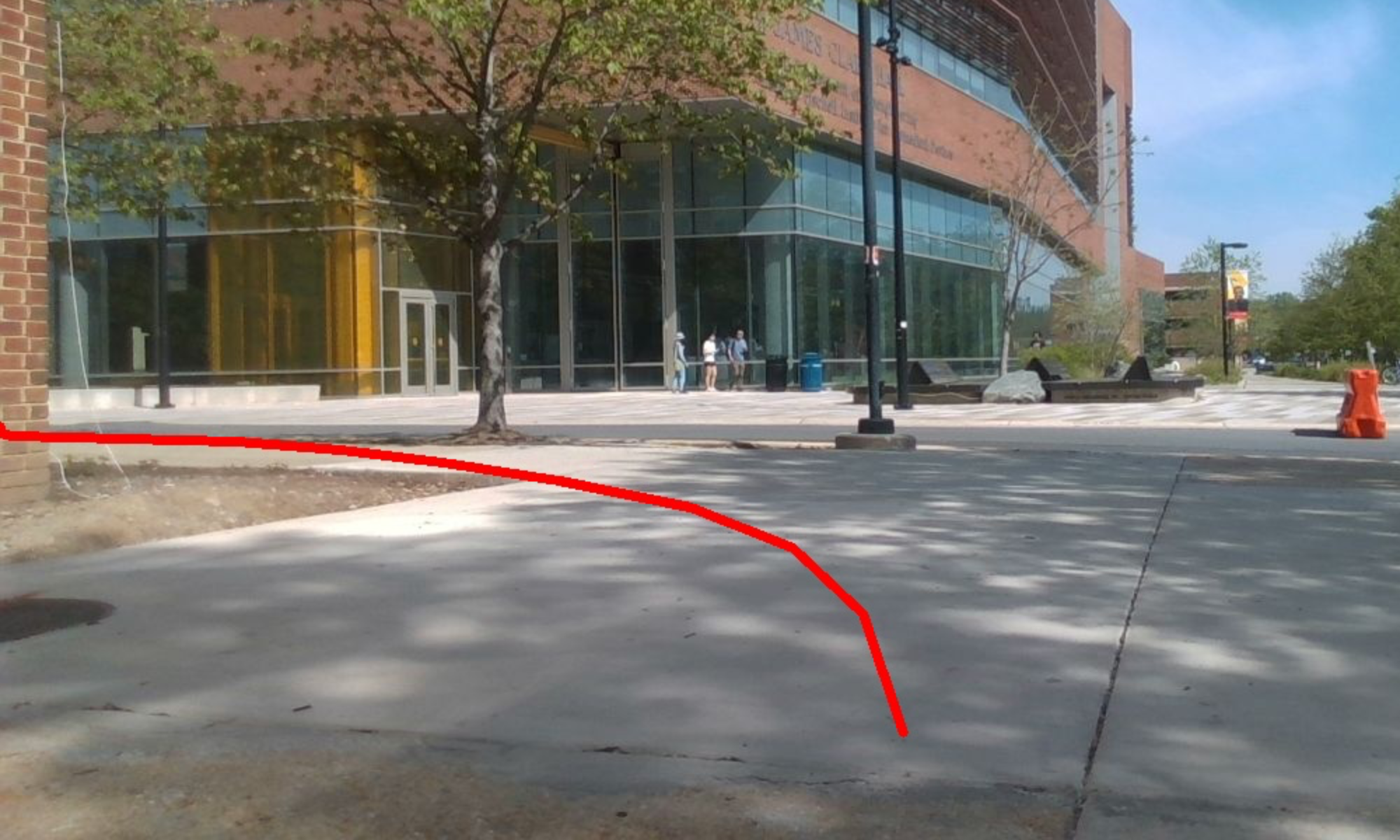} \end{minipage}
    &
    \begin{minipage}{.22\linewidth} \includegraphics[width=\linewidth]{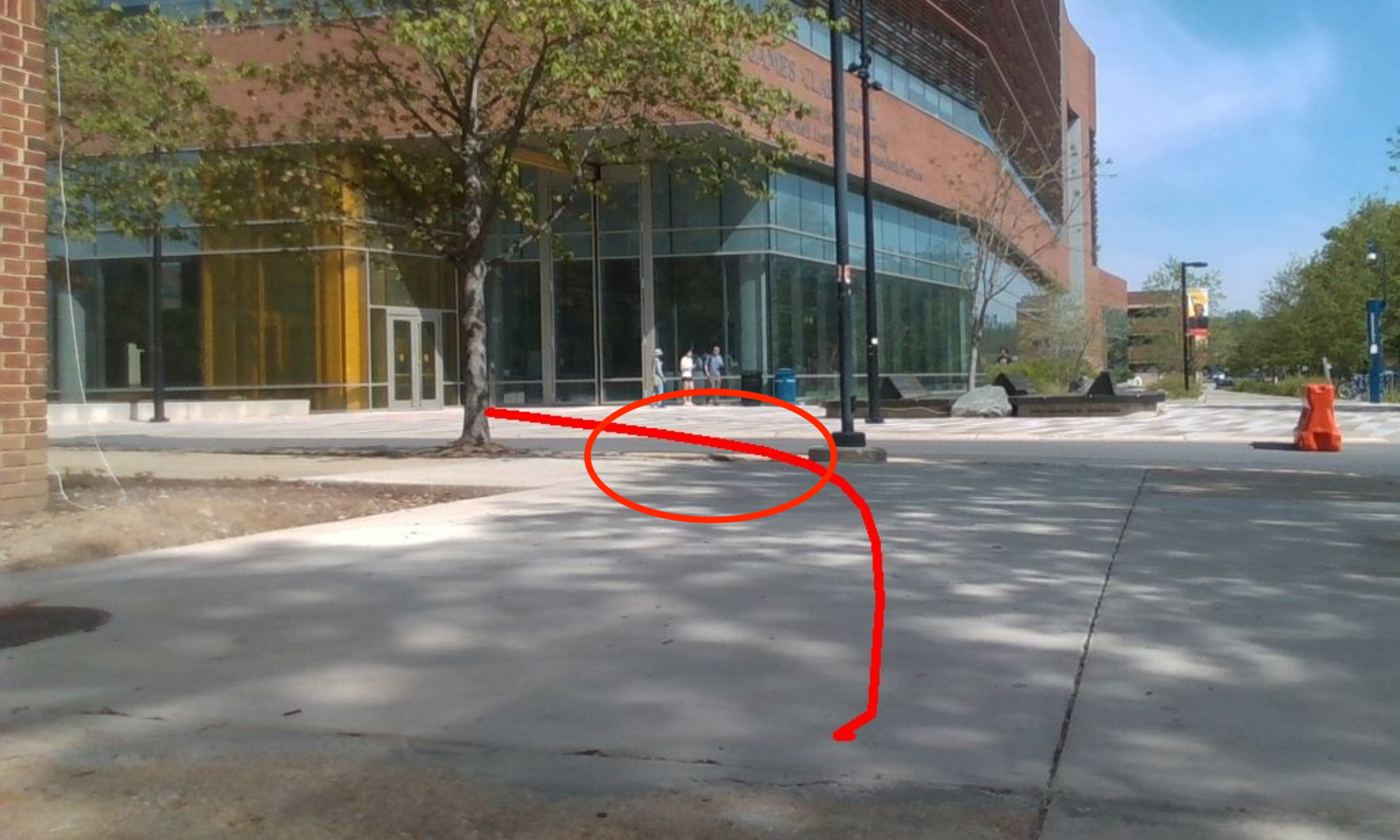} \end{minipage}
    &
    \begin{minipage}{.22\linewidth} \includegraphics[width=\linewidth]{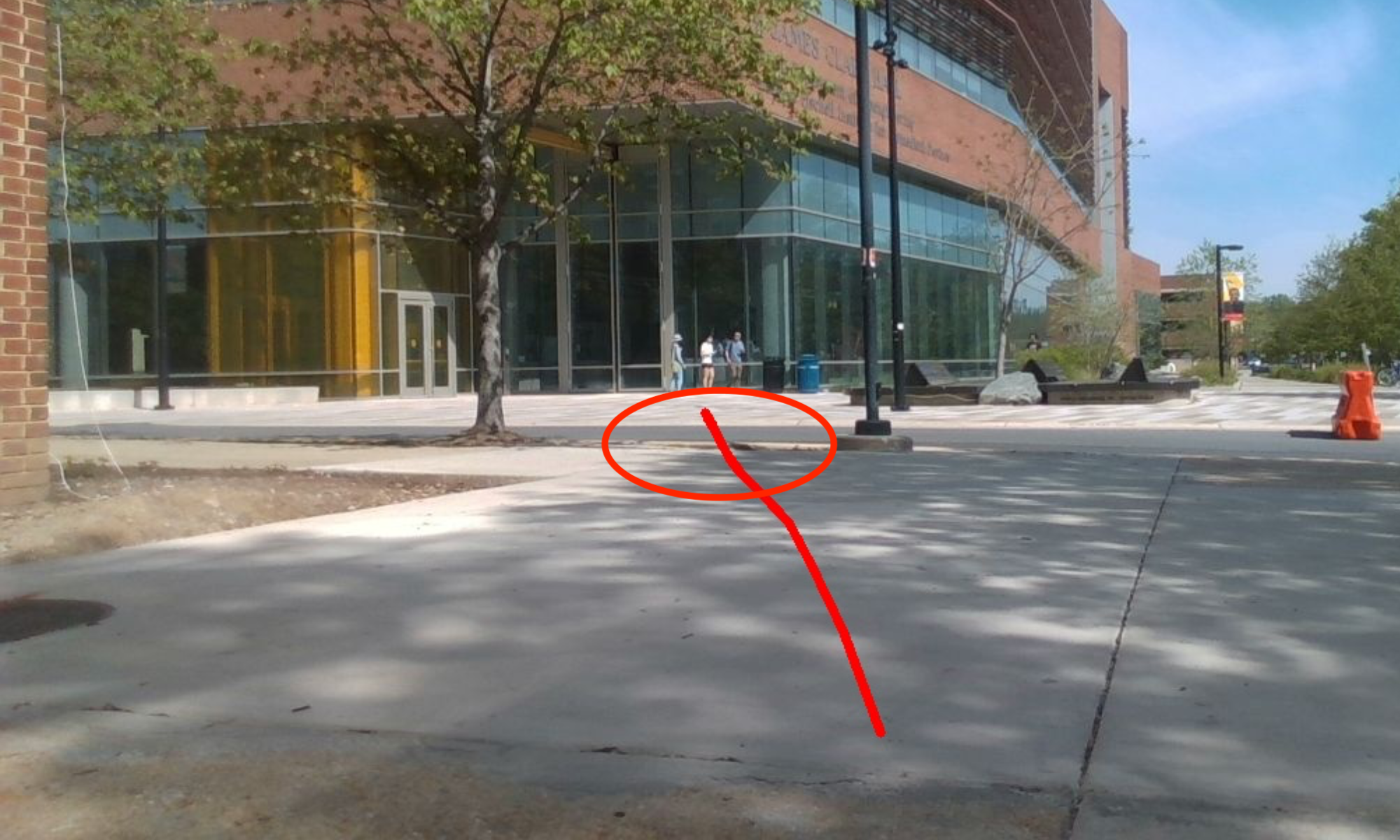} \end{minipage}
    &
    \begin{minipage}{.22\linewidth} \includegraphics[width=\linewidth]{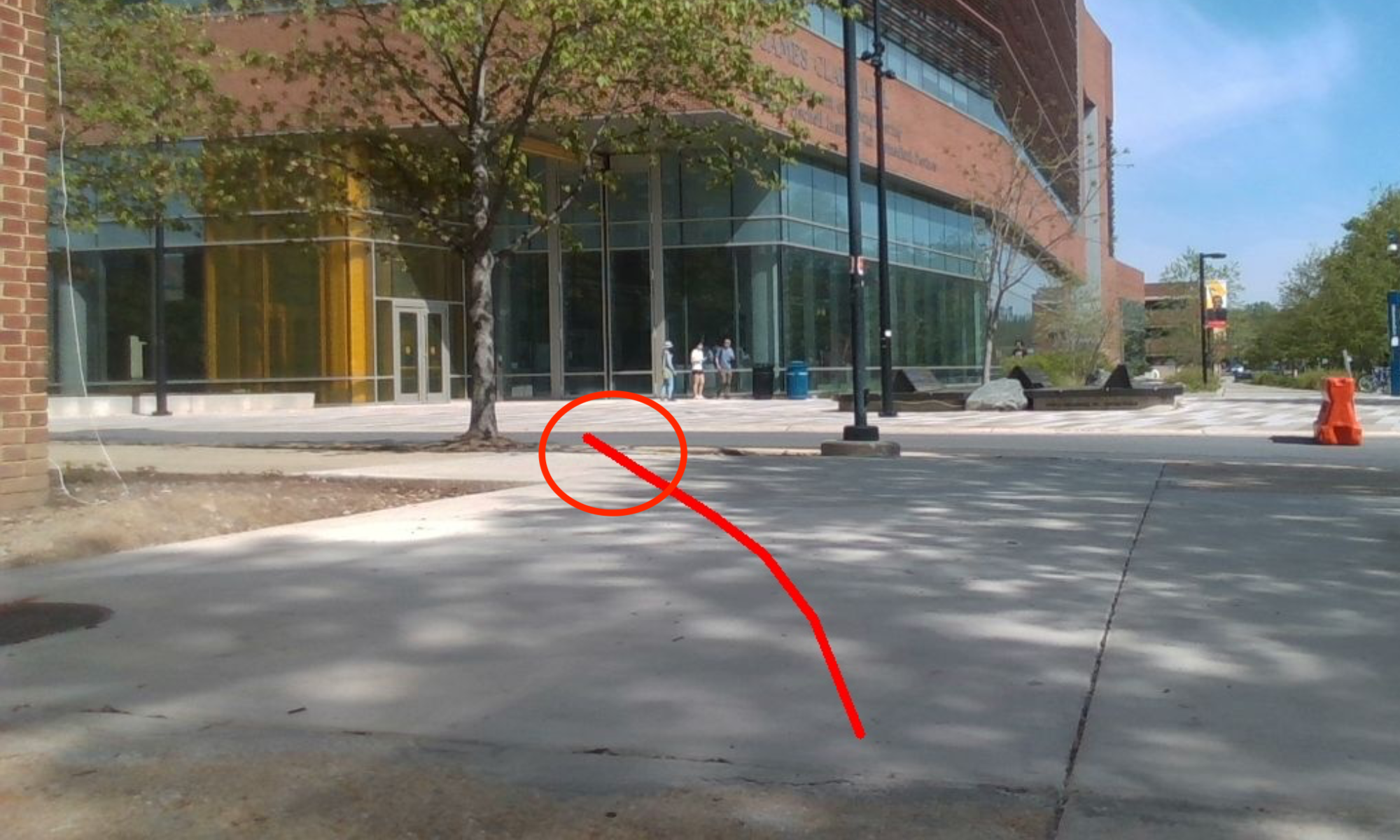} \end{minipage}
    \\ \hline

    \begin{minipage}{.22\linewidth} \includegraphics[width=\linewidth]{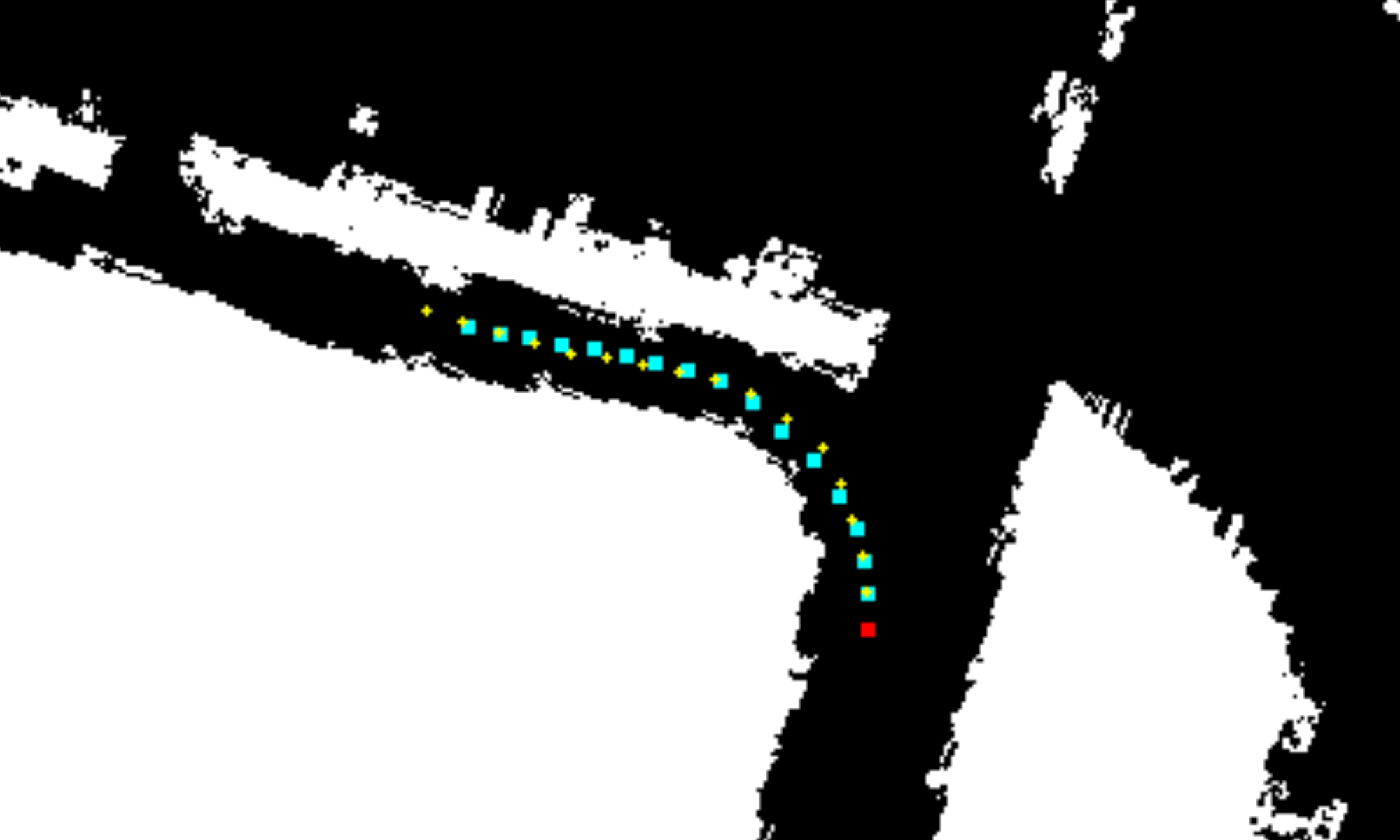} \end{minipage}
    &
    \begin{minipage}{.22\linewidth} \includegraphics[width=\linewidth]{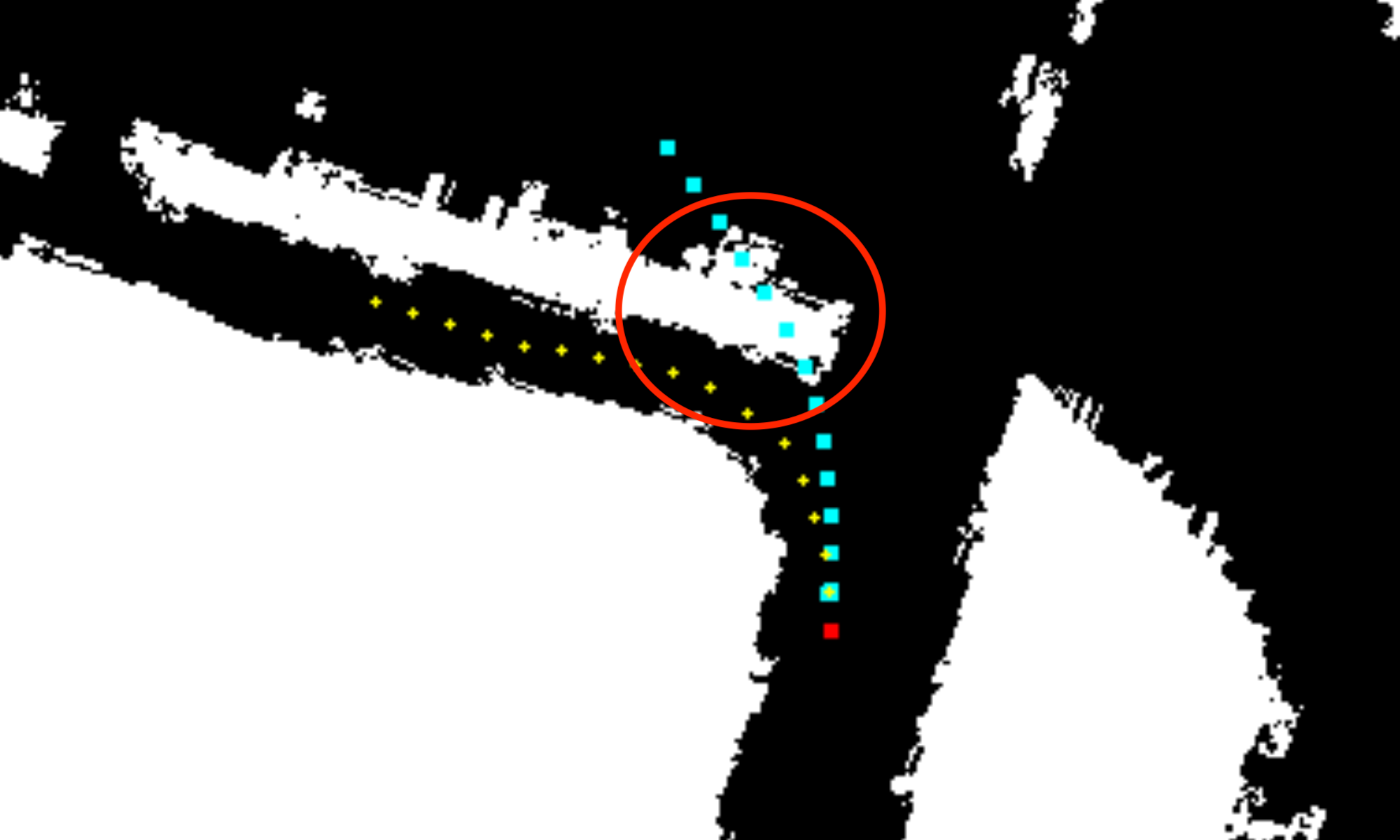} \end{minipage}
    &
    \begin{minipage}{.22\linewidth} \includegraphics[width=\linewidth]{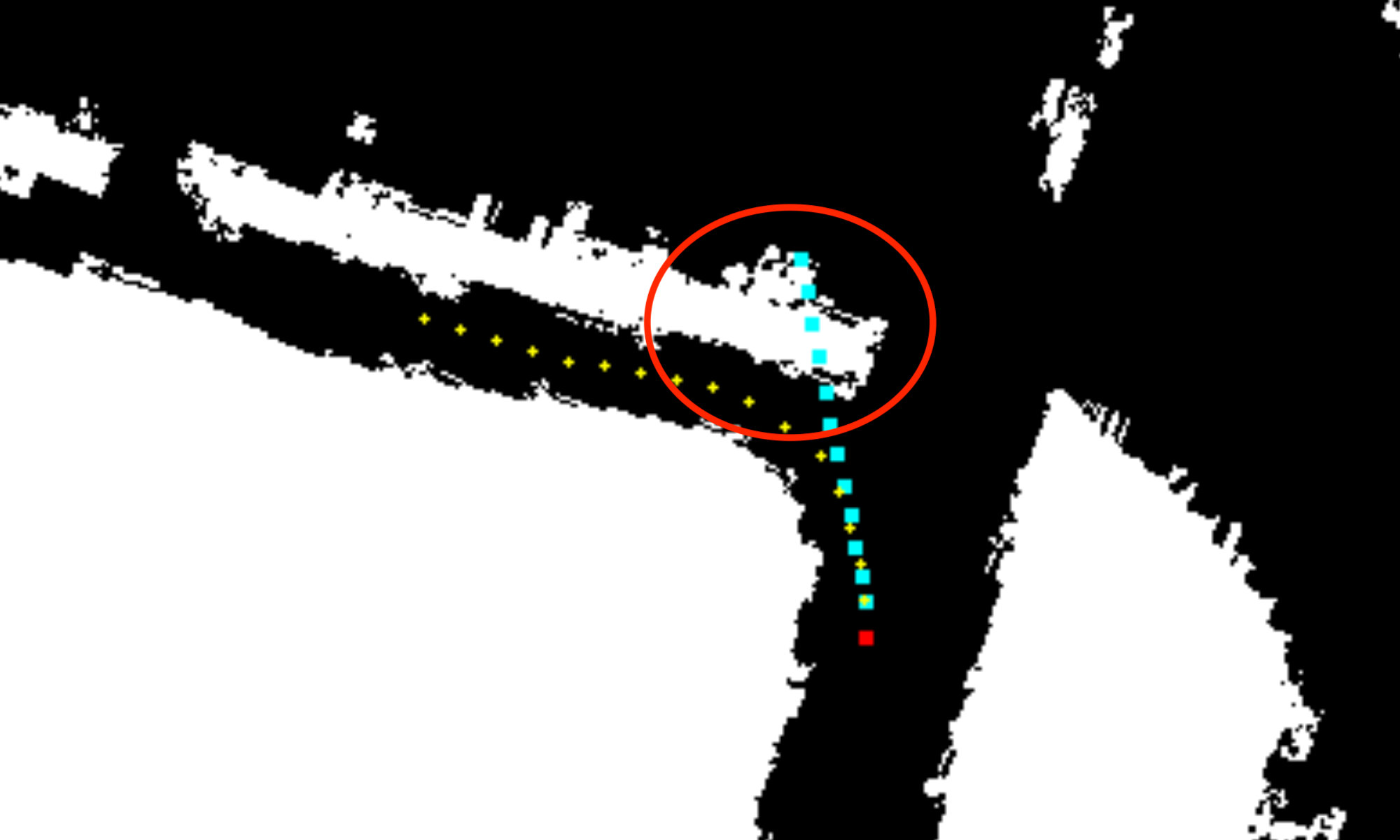} \end{minipage}
    &
    \begin{minipage}{.22\linewidth} \includegraphics[width=\linewidth]{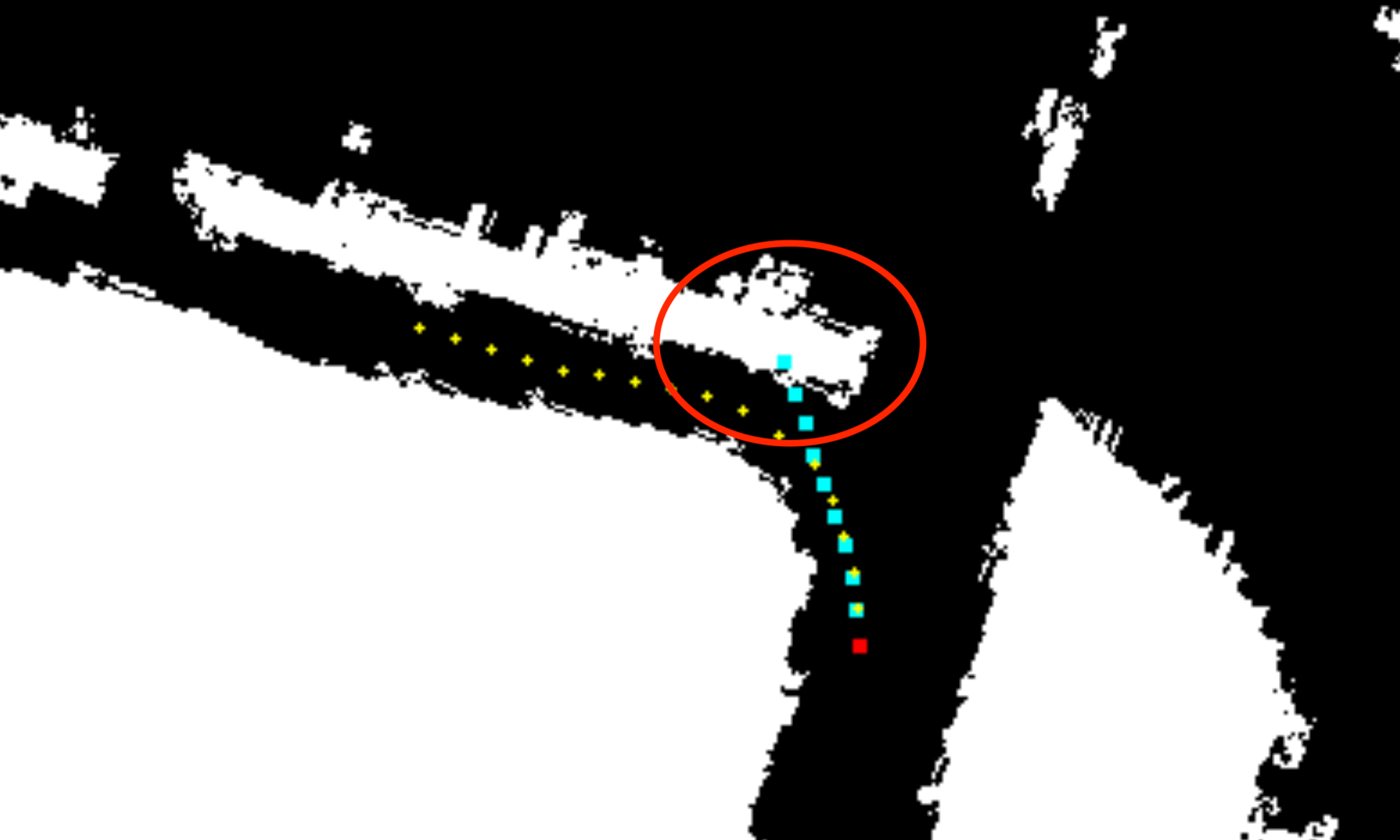} \end{minipage}

    \\ \hline
  \end{tabular}
  \caption{\textbf{Traversability Analysis in Challenging Occluded Environment: } The top row shows the generated trajectories (red) in the camera view. The bottom row shows the top-down view of the traversability map. The cyan color represents the generated trajectories, and the yellow color represents the most heuristic trajectory to the goal. DTG can generate trajectories w.r.t. the geometric shape of the traversable areas, but other approaches cannot generate fully traversable trajectories; the non-traversable parts are marked by red circles.}
  \label{fig:qualitative_traversability}
  \vspace{-1em}
\end{figure*}

\begin{figure*}[!ht]
  \centering
  \begin{tabular}{ | c | c | c | c |}
    \hline
   % \multicolumn{3}{|c}{\textbf{Traversability Analysis}} & \multicolumn{3}{|c|}{\textbf{Heuristic Analysis}} \\ \hline
   \textbf{DTG} & \textbf{MTG}~\cite{liang2023mtg} & \textbf{NoMaD}~\cite{sridhar2023nomad} & \textbf{ViNT}~\cite{shah2023vint} \\ \hline
   
    \begin{minipage}{.22\linewidth} \includegraphics[width=\linewidth]{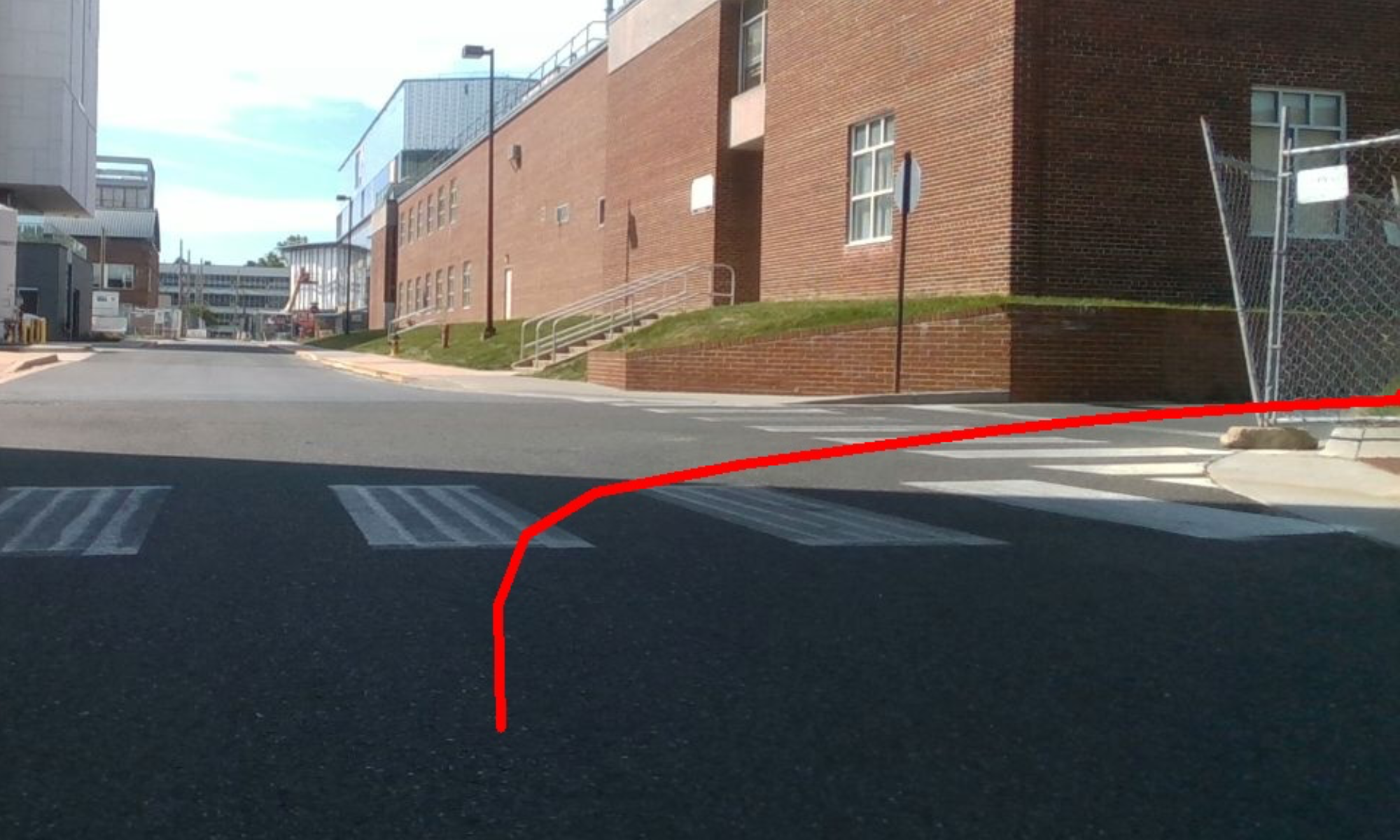} \end{minipage}
    &
    \begin{minipage}{.22\linewidth} \includegraphics[width=\linewidth]{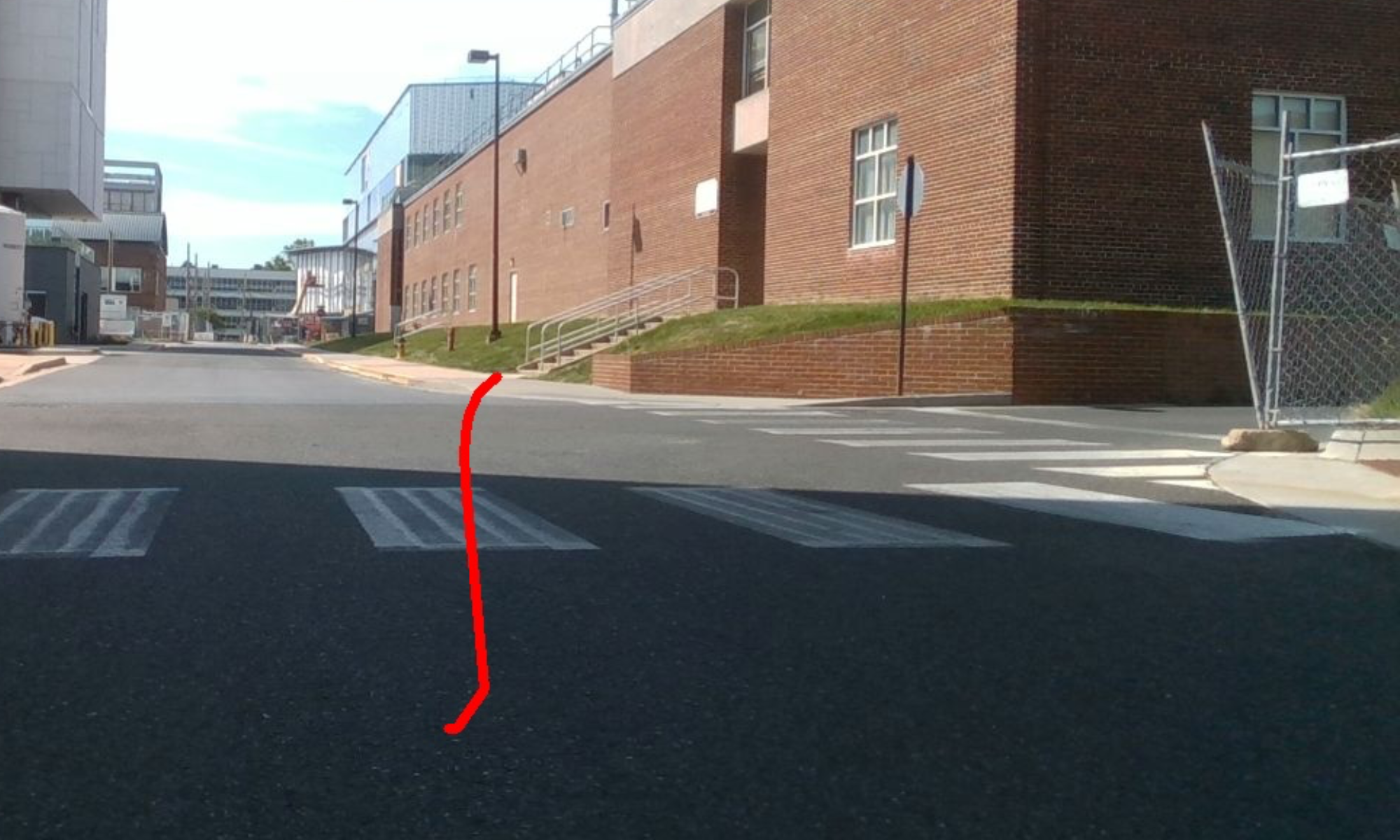} \end{minipage}
    &
    \begin{minipage}{.22\linewidth} \includegraphics[width=\linewidth]{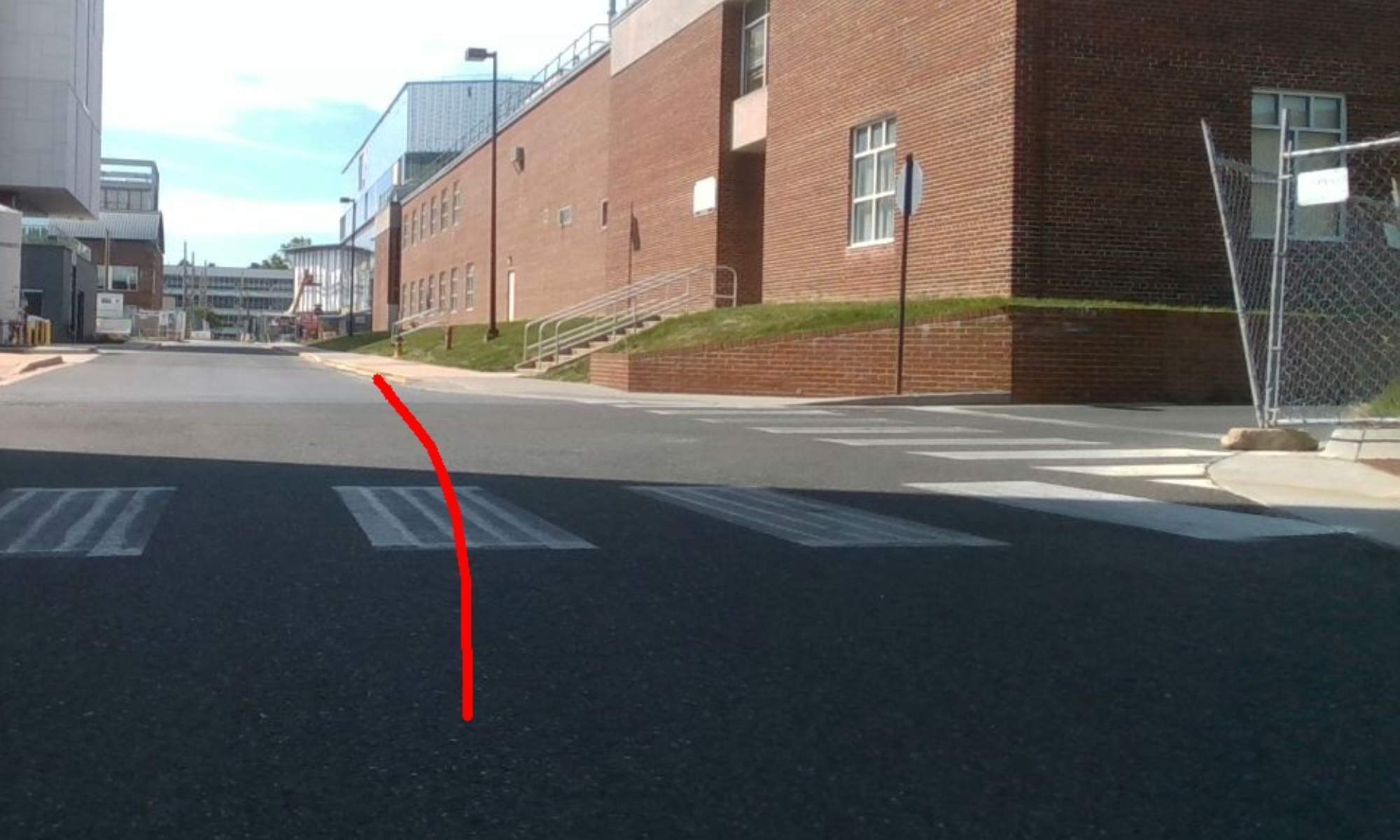} \end{minipage}
    &
    \begin{minipage}{.22\linewidth} \includegraphics[width=\linewidth]{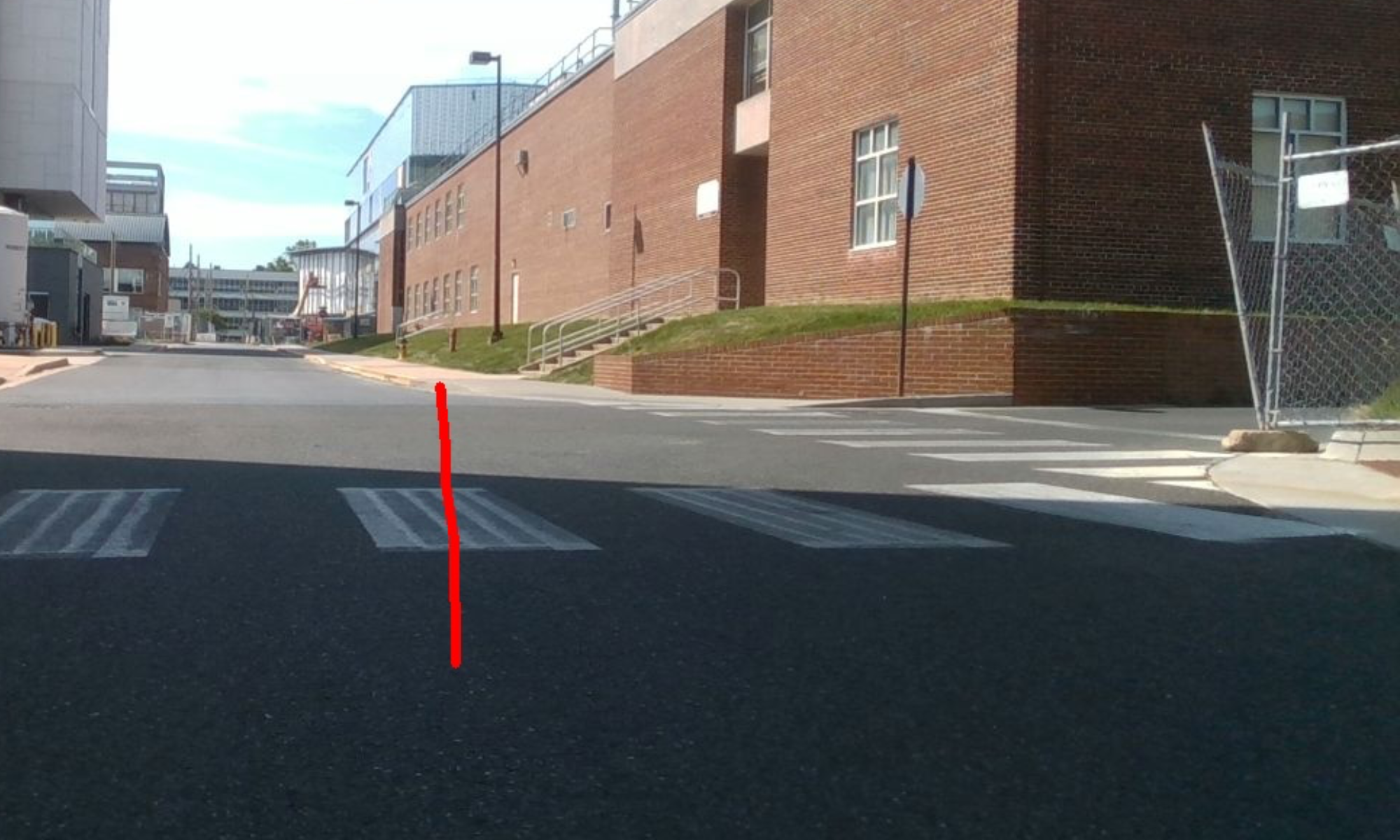} \end{minipage}
    \\ \hline
    \begin{minipage}{.22\linewidth} \includegraphics[width=\linewidth]{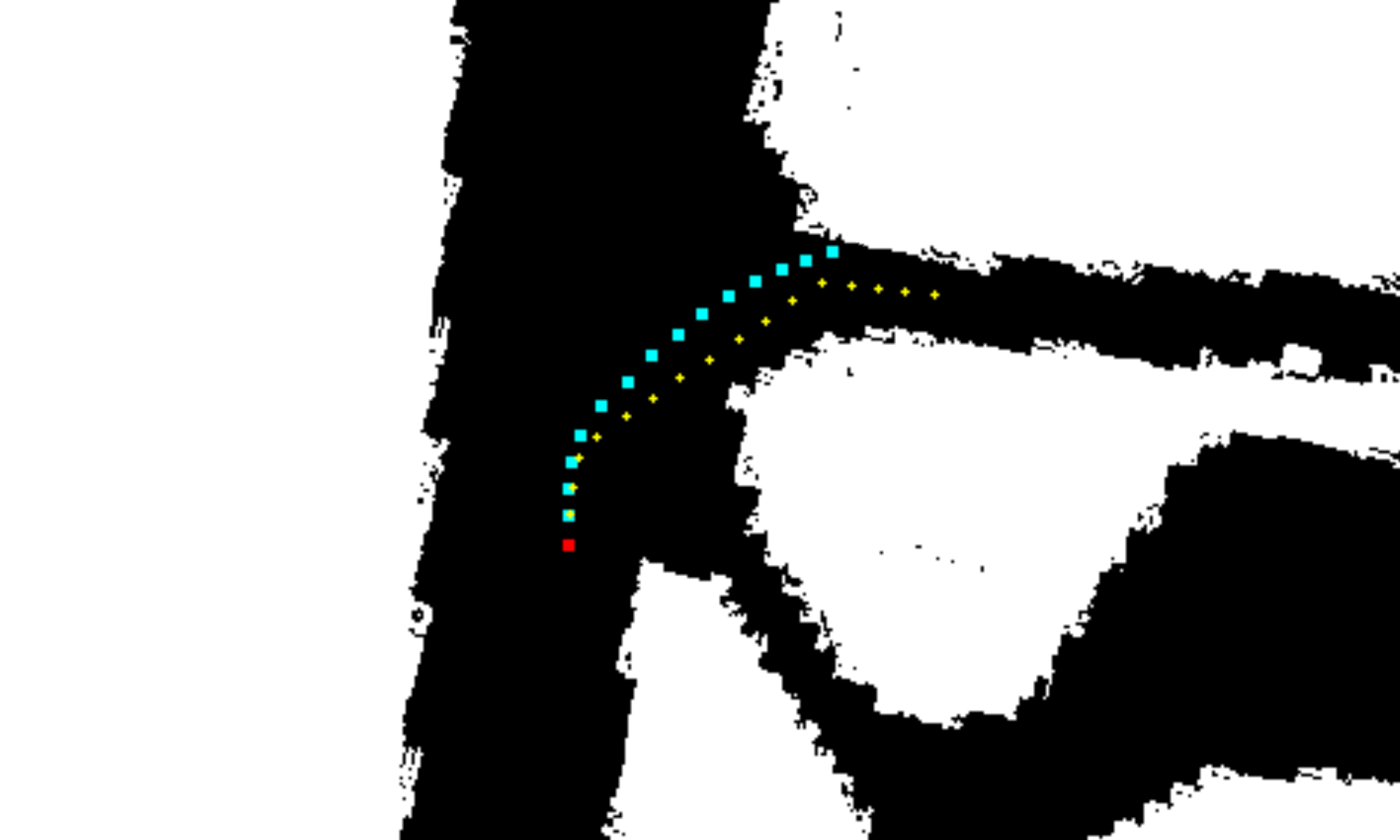} \end{minipage}
    &
    \begin{minipage}{.22\linewidth} \includegraphics[width=\linewidth]{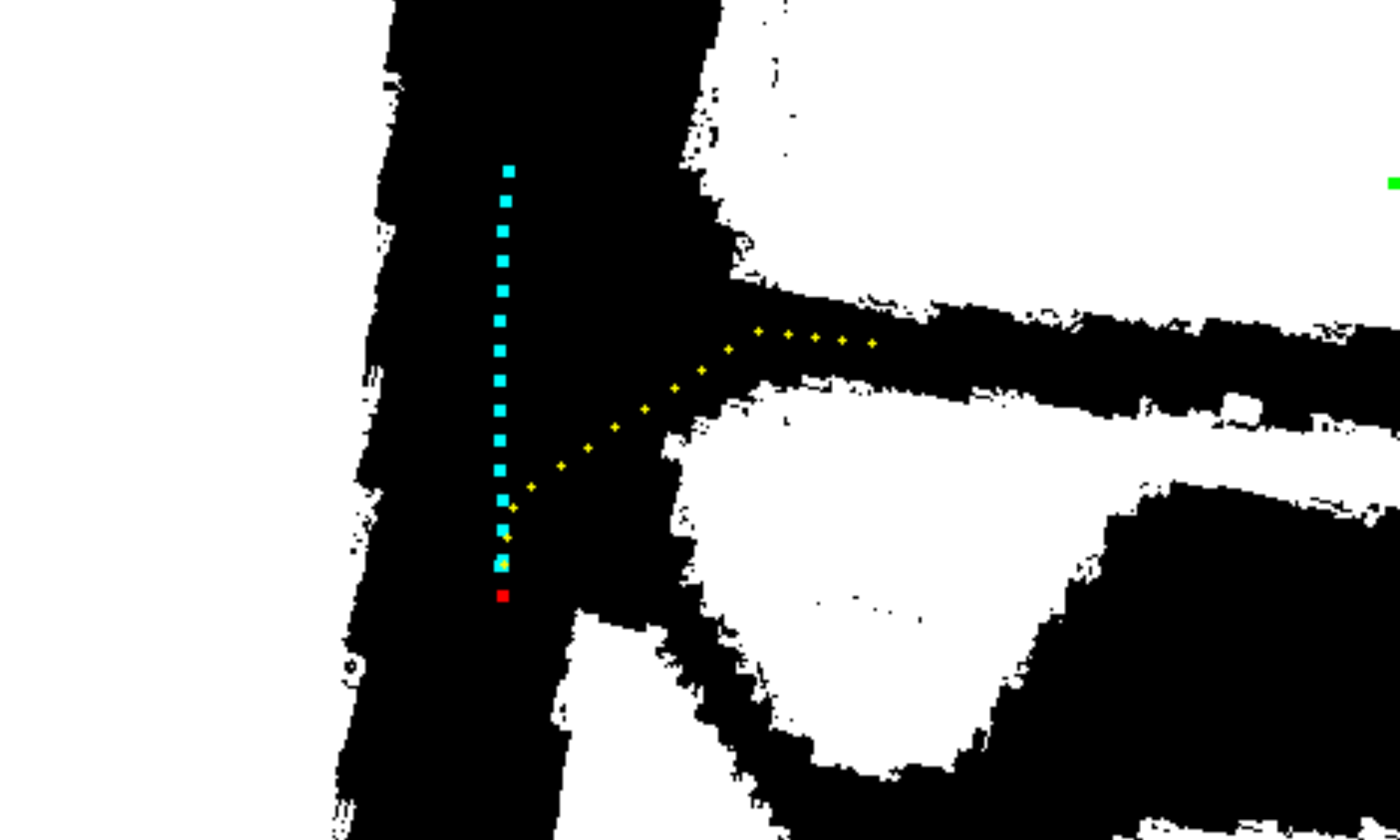} \end{minipage}
    &
    \begin{minipage}{.22\linewidth} \includegraphics[width=\linewidth]{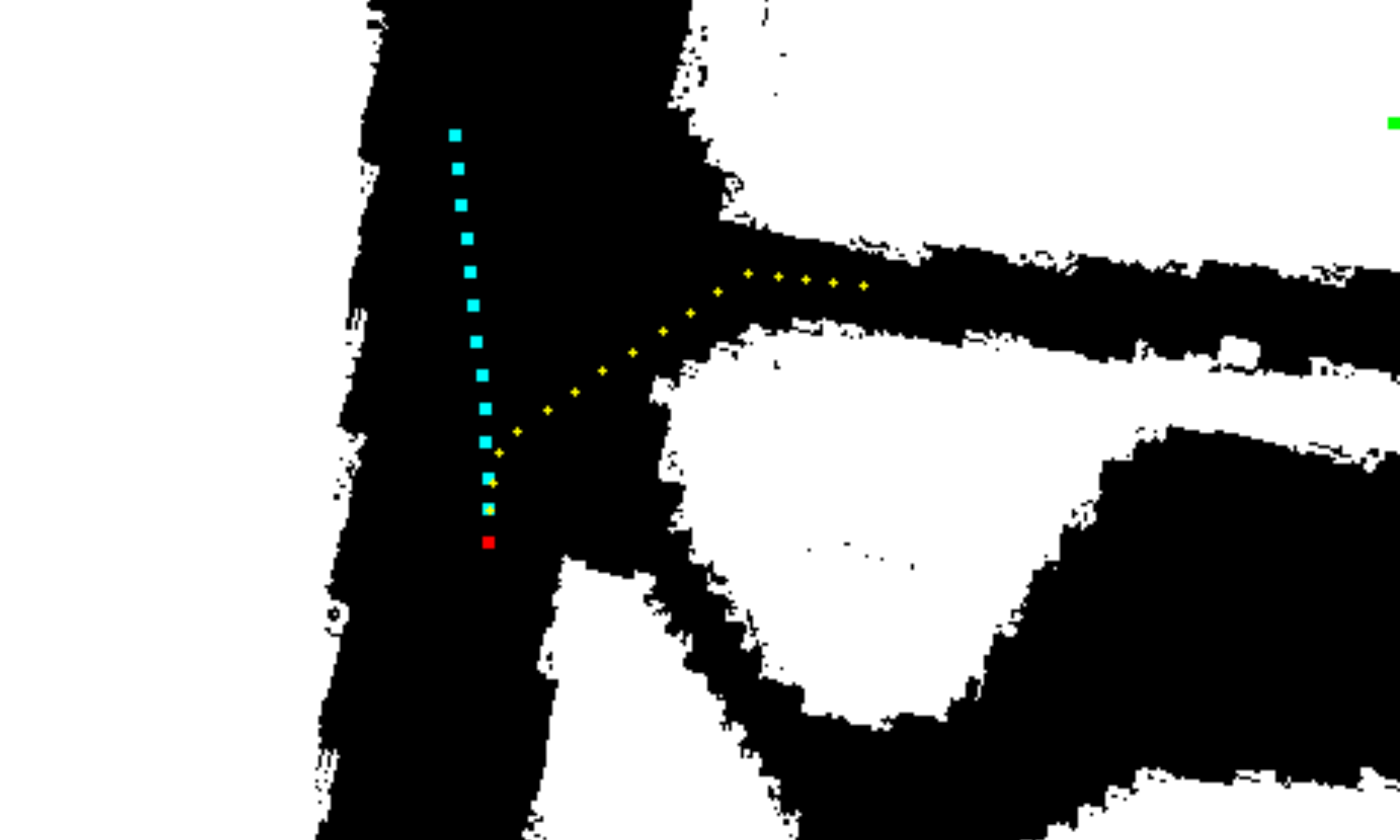} \end{minipage}
    &
    \begin{minipage}{.22\linewidth} \includegraphics[width=\linewidth]{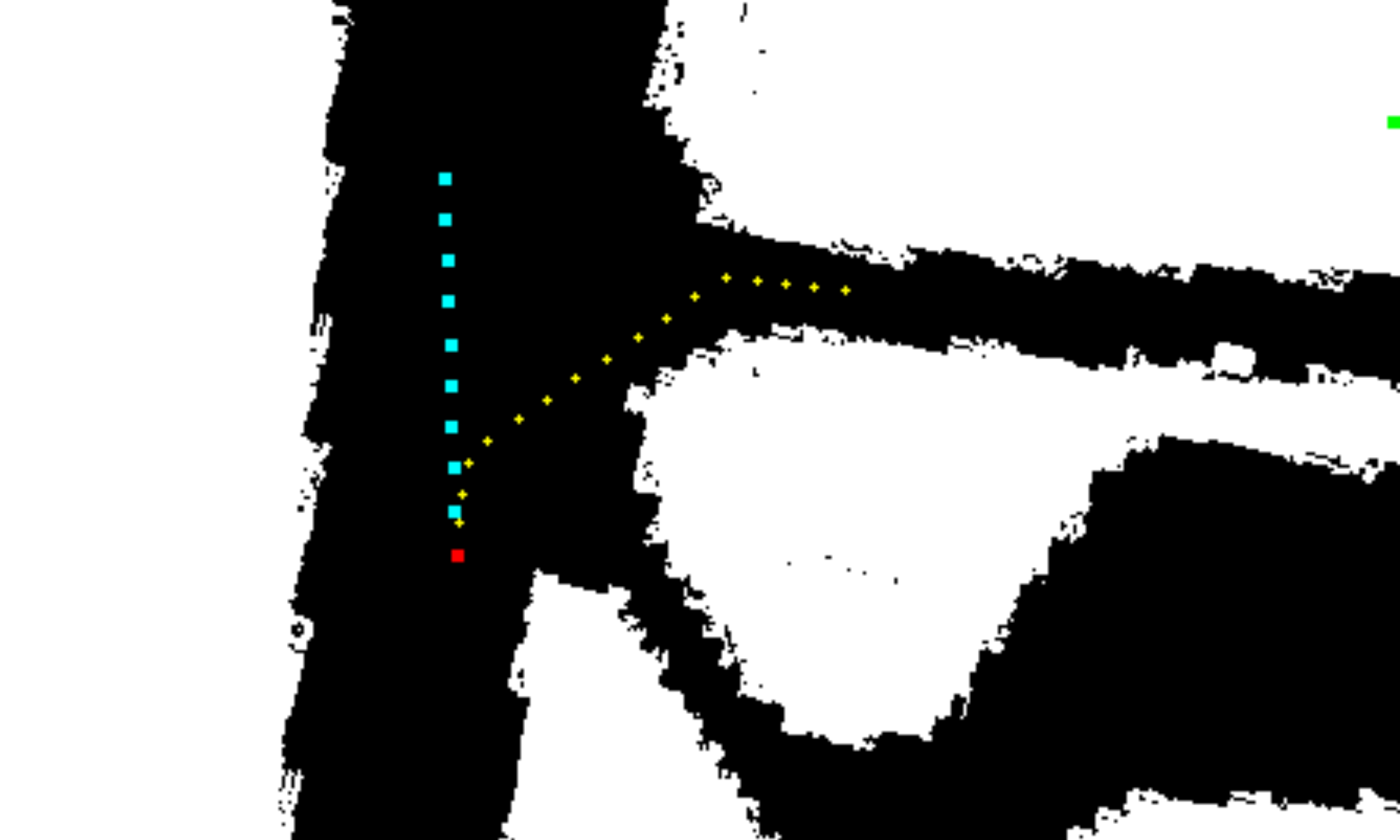} \end{minipage}
    \\ \hline
  \end{tabular}
  \caption{\textbf{Travel-Distance Analysis in Challenging Narrow Passage Environment: } The goal is behind the building. DTG generates a trajectory in a narrower space instead of the wide main road, leading to a shorter travel distance to the goal.}
  \label{fig:qualitative_distance}
\vspace{-2em}
\end{figure*}

\section{EXPERIMENTS}
\label{sec:experiments}

In this section, we discuss the details of the implementation, comparisons, and ablation studies of this approach. The experiments are designed to demonstrate the benefits of our innovations:  
\begin{enumerate}
    \item \textbf{Evaluate the novel end-to-end model, DTG, in maples outdoor global navigation:} We compare our approach, DTG, with SOTA outdoor navigation algorithms, including MTG~\cite{liang2023mtg}, NoMaD~\cite{sridhar2023nomad}, and ViNT~\cite{shah2023vint} in both a testing dataset and challenging outdoor real-world scenarios. The real-world experiment is achieved by combining the lower-level motion planner DWA~\cite{fox1997dynamic}. The details of the real-world experiment and the results are in Appendix~\ref{app:realworld_exp}\cite{supplements}.
    \item \textbf{Evaluate the efficacies of Diffusion model, CRNN diffusion cells, and adaptive traversability loss for training:} For each innovation, we conduct ablation studies to demonstrate the benefit and effect of the components. The diffusion mechanism is visualized in Appendix \ref{app:diffusion}\cite{supplements}
\end{enumerate}

\subsection{Implementation}
As described in Section~\ref{sec:approach}, our major perceptual sensor is a 3Hz 3D Velodyne Lidar (VLP-16) with 16 channels. The inputs of this approach contain $C_l=3$ consecutive frames of Lidar and $C_v=20$ consecutive frames of historic velocities. The goal is set by converting the GPS value to meters. During training, the goals are randomly selected within 60 meters; for testing, the goals are selected beyond 50 meters. The trajectory generator, DTG, generates $M=16$ waypoints in each trajectory.  The voxelization radius of the PointCNN in the Perception Encoder is 0.08m in this approach. The training data is the same as MTG~\cite{liang2023mtg}, which is collected by a Husky robot. \rev{The training and evaluation datasets are in different areas as shown in \cite{liang2023mtg}.} Velocities are collected from a 10Hz odometer. To keep all the perceptive information in the same time period, we select consecutive 10 frames of velocities as input. During training, the ground truth trajectories, generated by the A* algorithm with the shortest travel distance to the goal, are thresholded 15 meters from the robot’s position, and have 16 waypoints each. The training and evaluation are processed in a computer with an NVIDIA RTX A5000 GPU and an Intel Xeon(R) W-2255 CPU, and the real-world experiment is executed on a laptop with an Intel i7 CPU and one Nvidia GTX 1080 GPU. In the real-world experiment, the diffusion model generates trajectories in 5Hz. The network details of the architecture are in Appendix~\ref{app:details}\cite{supplements}.

\subsection{Evaluation}
In this section, we qualitatively and quantitatively evaluate DTG compared with different state-of-the-art methods and modified versions for ablation study. In the experiment, we compare DTG with ViNT~\cite{shah2023vint}, NoMaD~\cite{sridhar2023nomad}, and MTG~\cite{liang2023mtg}. Because NoMaD and ViNT require a sequence of images from the start position to the goal, we run the robot in the environment first and collect the images. Each consecutive pair of images has a distance of around 1 meter. MTG generates multiple trajectories, and we use the same method as the experiment in MTG~\cite{liang2023mtg} to choose the best trajectory. The evaluation metrics include: Traversability, Distance Ratio, Inference Time, and Model Size.

\textbf{Traversability: } Given a trajectory $\hat{\btau}$, the traversability is calculated as Equation \ref{eq:traversability}. The $c(\cdot)$ tells if the waypoint $\w_m$ is in the traversable area $\ca$. For K scenarios, we have the average traversability of the approach: $\frac{1}{K}\sum_{i=1}^K tr(\ca, \hat{\btau})$.
% , similar as the traversability loss in Equation~\ref{eq:traversability_loss}. Considering the robot has a radius of 0.5 meters, we clip the distance function to $[0-0.5]$ meters:
\begin{align}
    tr(\ca, \hat{\btau}) = \bigcap_{m=1}^M c(\ca, w_m), \;\; \w_m \in \hat{\btau}.
    % tr = \frac{1}{M}\sum_{m=1}^M\min{d(\Tilde{\ca}, w_m)}, \;\; w_m \in \hat{\tau}
    \label{eq:traversability}
\end{align}

\textbf{Distance Ratio: } The distance ratio is to evaluate the future travel distance of the trajectory to the goal and it measures the ratio of the travel distance of the trajectory w.r.t. the shortest travel distance from the robot's position to the goal. Therefore, given the trajectory length $\abs{\hat{\btau}}$, the travel distance $h_c$ from the robot's position, and the travel distance $h_t$ from the last waypoint $\w_M\in \hat{\btau}$, the distance ratio is defined as Equation \ref{eq:distance_ratio}. The ratio is trajectory length independent, so we can use it to compare trajectory generators with different trajectory lengths. 
\begin{align}
    hr(\hat{\btau}) = 1 - \frac{\abs{h_t - h_c}}{2\abs{\hat{\btau}}}
    \label{eq:distance_ratio}
\end{align}

\begin{table*}[]
    \centering
    % \vspace{-1em}
    \begin{tabular}{c c c c c c}
       \hline
        % \multirow{2}{*}{Evaluation} & Input & Distance   & Traversability & Inference & Model\\
        %  & Modality & Ratio ($\%$) & ($\%$)  &  Time (s) & Size (Mb) \\
         Evaluation & Input Modality & Distance Ratio ($\%$) & Traversability ($\%$) & Inference Time (s) & Model Size (Mb) \\
       \hline
       ViNT         & RGB & 66.02 & 79.02 & 0.69  & 113.49 \\ %\hline
       NoMaD        & RGB & 64.54 & 77.86 & 0.24  & 72.67 \\ %\hline
       MTG          & Lidar & 80.78 & 83.11 & 0.01  & 101.61 \\ \hline %\hline
       CVAE         & Lidar & 92.26 & 85.89 & 0.01  & 113.96 \\ %\hline
       DTG/t        & Lidar & 93.12 & 85.57 & 0.13  & 128.38 \\ %\hline
       DTG$_\text{U-Net}$ & Lidar & 92.94 & 90.74 & 2.09  & 1117.89 \\ %\hline
       DTG$_\text{CRNN}$  & Lidar & 93.61 & 89.00 & 0.13  & 128.38 \\ \hline
    \end{tabular}
    \caption{\textbf{Quantitative Results:} Our approach achieves  at least a $15\%$ improvement in distance ratio and around $7\%$ increase in traversability over other approaches. Our innovative components, CRNN, adaptive traversability training, and end-to-end diffusion-based generator also show effective improvement in global navigation task.}
    \label{tab:ros_quanti}
\vspace{-2em}
\end{table*}

\textbf{Comparisons: } This section shows the efficacy of DTG for mapless outdoor global navigation. From Table~\ref{tab:ros_quanti}, we observe our approach, DTG, outperforms other SOTA approaches. Specifically, we achieve by at least $15\%$ in distance ratio and $12.6\%$ in traversability improvement compared with ViNT and NoMaD. NoMaD and ViNT only take RGB images, so the robot is not very robust in generating trajectories only in traversable areas. NoMaD is better than ViNT w.r.t. inference time but slightly worse in traversability. Our approach outperforms MTG around $7\%$ in traversability and $16\%$ in distance ratio. The MTG has relatively good traversability but cannot choose the trajectory with the best distance ratio because in MTG the trajectories are generated by only comparing the straight distance to the goal instead of estimating the real travel distance of the trajectory. DTG also outperforms NoMaD and ViNT in inference time by 0.11s and 0.56s, but the model size is bigger than ViNT, NoMaD, and MTG. 

As shown in Figures~\ref{fig:qualitative_traversability} and \ref{fig:qualitative_distance}, we provide the qualitative explanation of how our approach outperforms other SOTA methods. The top row shows the generated trajectory in the camera and the bottom row shows the trajectory in the traversability map, where cyan represents the generated trajectory and yellow dots are the ground truth trajectories with the shortest travel distance to the goal. From the two figures, we observe NoMaD and ViNT generate shorter trajectories than MTG or DTG. Figure \ref{fig:qualitative_traversability} shows a challenging occluded environment around a corner; DTG can generate the trajectory aligned with the geometric shape of the traversable areas, but MTG and NoMaD do not perform well in challenging corner situations. In Figure~\ref{fig:qualitative_distance}, the goal is behind the building and the ground truth trajectory lies in a narrow passage. The scenario is challenging in terms of travel distance estimation. DTG can still generate a trajectory to the target in the narrow passage, while other approaches all choose trajectories in the wider main road. 

\textbf{Ablation Study: } To evaluate the capability of different components of our innovations, we compare DTG with the modified versions without traversability loss during training, changing our Conditional RNN to the regular U-Net model and changing the generative model from the diffusion model to CVAE~\cite{cvae}. From Table \ref{tab:ros_quanti}, our DTG has the best heuristic compared with other ablation studies. The U-Net is much larger than our proposed CRNN with more than 989.51Mb and is also much slower than all other approaches, but because the U-Net model is larger than CRNN, it can encode information better and generates trajectories with slightly better traversability. Our novel model CRNN is faster and smaller than U-Net, but we achieve comparable distance ratios and traversability in trajectory generation. The DTG/t shows the model without training traversability loss. Obviously, it has the worst traversability. The CVAE model is smaller, but the generative capability is not as good as diffusion models. Its distance ratio and traversability of the generated trajectories are worse than DTG.

\section{Conclusion, Limitations, and Future Work}
\label{sec:conclusion}
We present a novel end-to-end diffusion-based trajectory generator for mapless global navigation and demonstrate the innovations of the end-to-end approach and the efficacy of the different innovative components in both evaluation dataset and the real-world experiments. We achieve at least a $15\%$ improvement in distance ratio and a $7\%$ improvement in traversability over other SOTA approaches.

There are also limitations of the approach, DTG. Because the trajectory generator generates trajectories in real-time, there should be some mechanism to smartly choose the best trajectory during navigation, e.g., estimating the feasibility and confidence of the trajectory.

% \noindent\textbf{Acknowledgment:} This work was supported in part by ARO Grants  W911NF2310046, W911NF2310352 and U.S. Army Cooperative Agreement W911NF2120076
\section*{ACKNOWLEDGMENT}
This work was supported in part by ARO Grants  W911NF2310046, W911NF2310352 and U.S. Army Cooperative Agreement W911NF2120076

\bibliographystyle{IEEEtran}
\bibliography{IEEEabrv,ref}

\clearpage
\newpage
\section{Appendix}
\label{appdenix}

\subsection{Real-world Experiments}
\begin{figure}
    \centering
    \includegraphics[width=\linewidth]{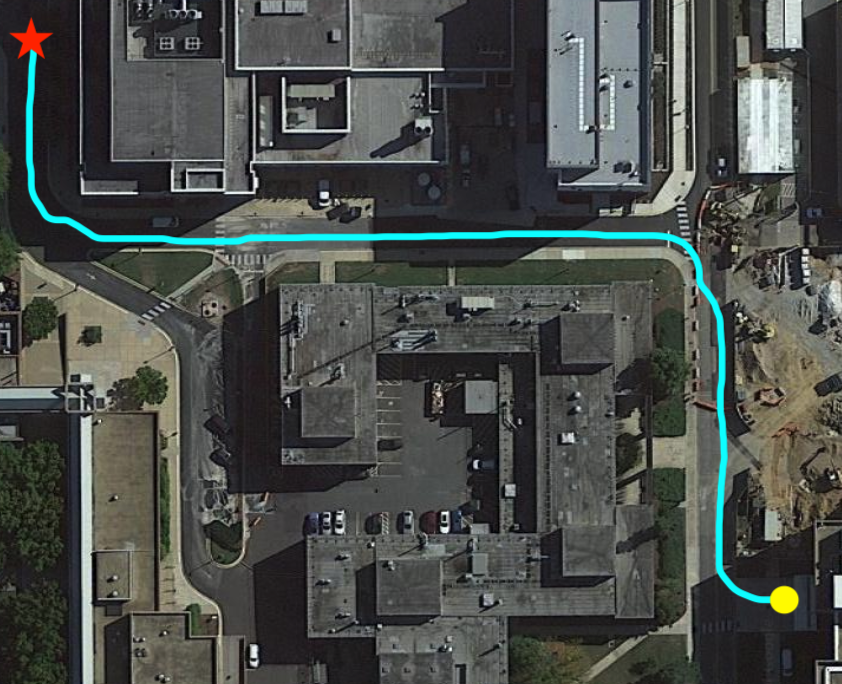}
    \caption{The red star indicates the start position and the yellow circle indicates the goal. The example trajectory from the start to the goal is the blue path, with around a 200-meter travel distance.}
    \label{fig:experiment_scenario}
\end{figure}

The real-world experiment is implemented with a Husky robot, which has the biggest velocity of 1m/s. The real-world experiment is applied in the scenarios with curbs, grass, buildings, etc. Global navigation requires long-range navigation, and in our experiment, the traveling distance between the start and the goal positions is around 200 meters, and the environment is as shown in Figure~\ref{fig:experiment_scenario}. The robot uses GPS to localize and detect if it arrives at the goal, which is within a radius of 20 meters from the goal GPS, considering the accuracy of the GPS device is around 20 meters around buildings. The pipeline contains two parts: a trajectory generator and a motion planner. Trajectory generators take observation data and output a trajectory and the motion planner, DWA~\cite{fox1997dynamic}, handles low-level collision avoidance and follows the waypoints from the generated trajectories. In the experiment, we fine-tune the trajectory publishing frequency of each algorithm to make them perform the best. We realize that our method doesn't require a high frequency of publishing trajectories because our trajectory is longer and has better quality w.r.t. the traversability, as shown in Video~\cite{supplements} as qualitative results. We also quantitatively analyze the results in Table \ref{tab:rw_quanti}. We calculate the travel distance until the robot achieves the goal or loses the traction of the topological map for NoMad and ViNT. The Human Interferes counts the times of human interaction with the robot when the robot is in collision or runs into non-traversable areas. We observe that DTG and MTG can reach the goal, but NoMaD and ViNT easily lose track of the topological map in the turning scenarios. Compared with MTG, our approach, DTG, has fewer human interferes and a shorter distance to the goal. 

\label{app:realworld_exp}
\begin{table}[]
    \centering
    \begin{tabular}{c|c|c|c|c}
       \hline
       Real-world & Input  & Traveling & Human & Reached \\ 
       Experiment & Modality & Distance & Interferes (/10) & Goal\\ \hline    
       ViNT       & RGB &66.61  & 0.2  & No \\ \hline
       NoMaD      & RGB & 80.18  & 0.2  & No\\ \hline
       MTG        & Lidar & 240.73  & 0.7   & Yes\\ \hline
       DTG        & Lidar & 212.91  & 0.2   & Yes\\ \hline
    \end{tabular}
    \caption{DTG and MTG achieves the goal but NoMaD and ViNT cannot. Our approach, DTG, has less traveling distance and also fewer human interferes.}
    \label{tab:rw_quanti}
\end{table}

\subsection{Diffusion Mechanism}
\label{app:diffusion}
\begin{figure}
    \centering
    \includegraphics[width=\linewidth]{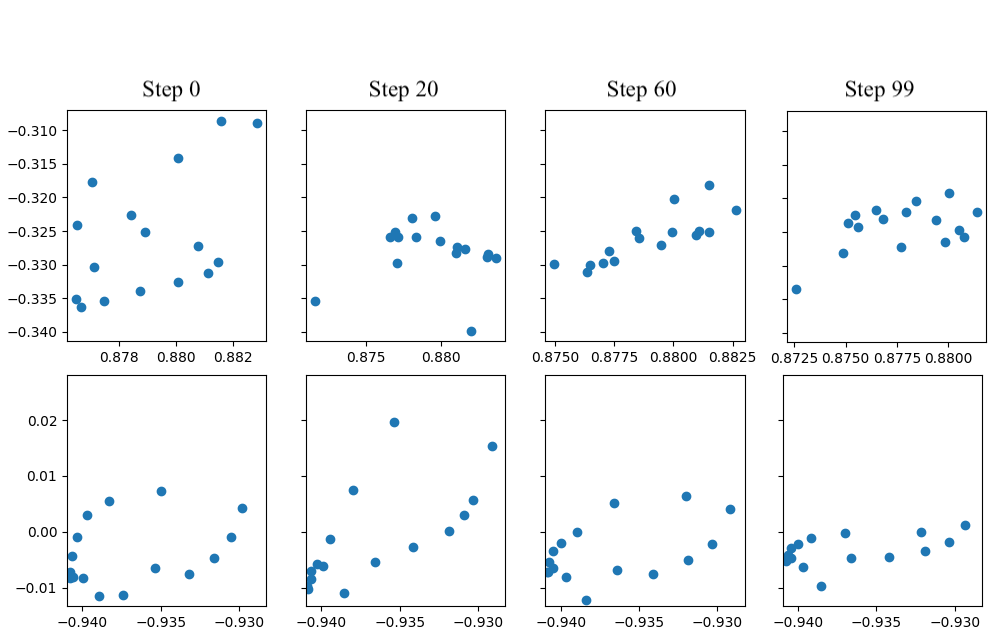}
    \caption{Diffusion Steps: In each generated trajectory, there are 16 waypoints. This figure shows the output values of the diffusion model; those values are distance increments for each waypoint (not waypoint positions). In the beginning steps, the output contains lots of noise; in later steps, the diffusion model denoises the increments, and those values are in a more concentrated area and reasonable shape.}
    \label{fig:diffusion_steps}
\end{figure}

To visualize the mechanism of the diffusion model in DTG, we show the output of the diffusion models $\set{\Delta x_m, \Delta y_m}$ in Figure \ref{fig:diffusion_steps}. In Step 0, the increment distances are random, but after several steps, they become more concentrated and form proper shapes.

\subsection{Details of the Architecture}
\label{app:details}

\begin{figure}
    \centering
    \includegraphics[width=\linewidth]{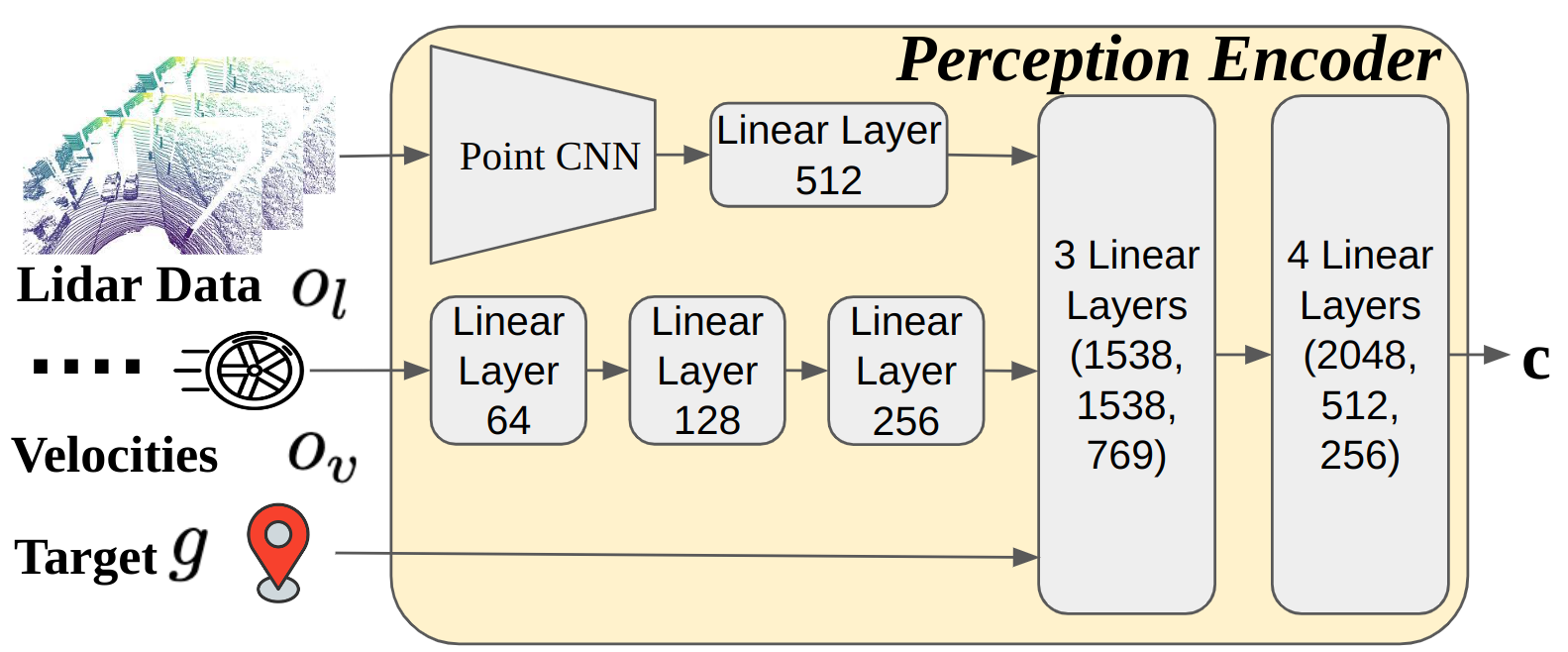}
    \caption{The details of the Perception Encoder.}
    \label{fig:perception_encoder}
\end{figure}

The details of the Perception Encoder are shown as Figure~\ref{fig:perception_encoder}.

\subsection{Failure Cases}
\begin{figure}
    \centering
    \includegraphics[width=\linewidth]{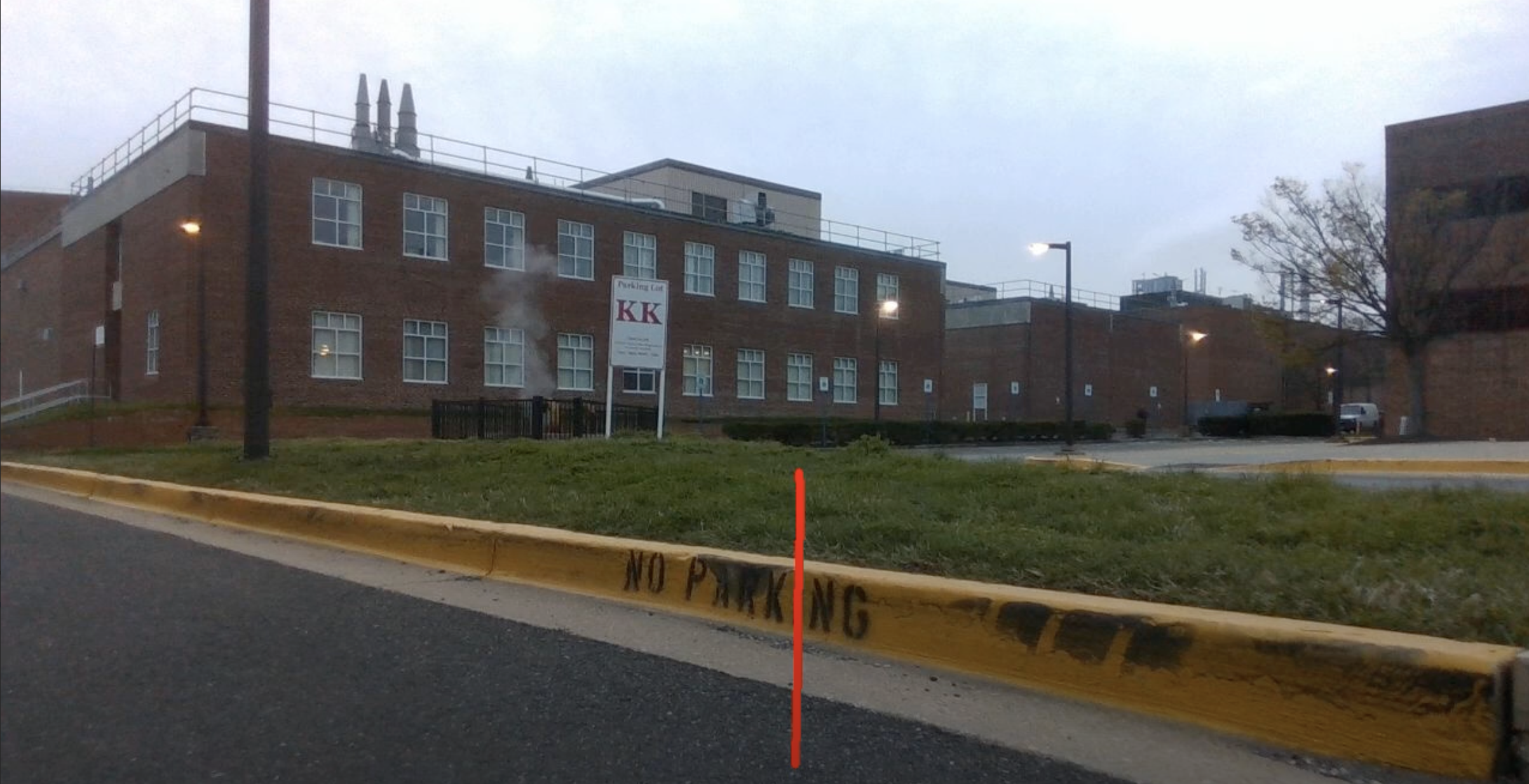}
    \caption{Failure Case: the model confuses when the both sides of curbs have similar flatness.}
    \label{fig:failure_cases}
\end{figure}

Considering the Lidar point clouds are sparse and because of the data quality that the curbs are not well segmented, the model is not good at detecting the curbs where both sides of the curbs have very similar flatness and the generated trajectories may lie on the curbs as Figure \ref{fig:failure_cases}.
\end{document}